\documentclass[preprint,11pt]{elsarticle}

\usepackage{amssymb}

\journal{Knowledge-Based Systems}

\usepackage{hyperref}
\usepackage{amsmath}
\usepackage{amsfonts}
\usepackage{mathrsfs}
\usepackage{algorithmic}
\usepackage{graphicx}
\usepackage{textcomp}
\usepackage{subfig}
\usepackage{soul}
\usepackage{ulem}
\usepackage{caption} 
\usepackage{colortbl}
\usepackage{cancel}
\usepackage{ulem}
\usepackage{xspace}
\usepackage{makecell}
\usepackage{tikz}
\usepackage{multirow}
\usepackage{scalerel}
\usepackage{xcolor}

\def\mcirc{\mathbin{\scalerel*{\bigcirc}{t}}}

\definecolor{MediumVioletRed}{RGB}{199, 21, 133}
\definecolor{Coral}{RGB}{255, 127, 80}
\definecolor{CornflowerBlue}{RGB}{100, 149, 237}

\newcommand{\indomain}{\raisebox{3pt}{\footnotesize\colorbox{MediumVioletRed!80}{\ \ \ }}\xspace}
\newcommand{\generaldomain}{\raisebox{3pt}{\footnotesize\colorbox{CornflowerBlue!80}{\ \ \ }}\xspace}
\newcommand{\mixeddomain}{\raisebox{3pt}{\footnotesize\colorbox{Coral!80}{\ \ \ }}\xspace}

\newcommand{\indomaincolor}{violet\xspace}
\newcommand{\generaldomaincolor}{blue\xspace}
\newcommand{\mixeddomaincolor}{coral\xspace}

\newcommand{\autoencodershape}{$\mcirc$\xspace}
\newcommand{\autoregressiveshape}{$\triangle$\xspace}
\newcommand{\texttotextshape}{$\square$\xspace}

\newcommand{\bert}{BERT\xspace}
\newcommand{\spanbert}{SpanBERT\xspace}
\newcommand{\roberta}{RoBERTa\xspace}
\newcommand{\electra}{ELECTRA\xspace}
\newcommand{\biobert}{BioBERT\xspace}
\newcommand{\bioclinicalbert}{BioClinicalBERT\xspace}
\newcommand{\scibert}{SciBERT\xspace}
\newcommand{\xlnet}{XLNet\xspace}
\newcommand{\pubmedbert}{PubMedBERT\xspace}
\newcommand{\bioroberta}{BioRoBERTa\xspace}
\newcommand{\bioelectra}{BioELECTRA\xspace}
\newcommand{\bertweet}{BERTweet\xspace}

\newcommand{\distilbert}{DistilBERT\xspace}
\newcommand{\xyendr}{EnDR-BERT\xspace}
\newcommand{\medgpttwo}{Med-GPT2\xspace}

\newcommand{\tfive}{T5\xspace}
\newcommand{\pegasus}{PEGASUS\xspace}
\newcommand{\gpttwo}{GPT-2\xspace}
\newcommand{\bart}{BART\xspace}
\newcommand{\scifive}{SciFive\xspace}

\newcommand{\autoencoder}{AutoEncoding\xspace}
\newcommand{\autoencoders}{AutoEncoding models\xspace}
\newcommand{\autoregressive}{AutoRegressive\xspace}
\newcommand{\texttotext}{Text-to-Text\xspace}
\newcommand{\shapvals}{Shapley values\xspace}

\newcommand{\crf}[1]{#1+ CRF}
\newcommand{\lstm}[1]{#1+ LSTM}

\newcommand{\modeltitle}[1]{~\newline\textbf{#1}}

\newcommand{\asd}[2]{#1 {\scriptsize $\pm$ #2}}

\expandafter\def\expandafter\UrlBreaks\expandafter{\UrlBreaks
\do\a\do\b\do\c\do\d\do\e\do\f\do\g\do\h\do\i\do\j%
\do\k\do\l\do\m\do\n\do\o\do\p\do\q\do\r\do\s\do\t%
\do\u\do\v\do\w\do\x\do\y\do\z\do\A\do\B\do\C\do\D%
\do\E\do\F\do\G\do\H\do\I\do\J\do\K\do\L\do\M\do\N%
\do\O\do\P\do\Q\do\R\do\S\do\T\do\U\do\V\do\W\do\X%
\do\Y\do\Z\do\-%
\do1\do2\do3\do4\do5\do6\do7\do8\do9\do0}

\newcommand{\specialcell}[2][c]{\begin{tabular}[#1]{@{}c@{}}#2\end{tabular}}
\newcommand{\quot}[1]{``#1''}

\newcommand{\TITLE}{Extensive Evaluation of Transformer-based Architectures for Adverse Drug Events Extraction}

\begin{document}

\begin{frontmatter}



\title{\TITLE\tnoteref{doi}}

\tnotetext[doi]{Published version: \url{https://doi.org/10.1016/j.knosys.2023.110675}}


\author[uniud]{Simone Scaboro}
\ead{scaboro.simone@spes.uniud.it}

\author[uniud,unina]{Beatrice Portelli\corref{cor1}}
\ead{portelli.beatrice@spes.uniud.it}

\author[polyu]{Emmanuele Chersoni}
\ead{emmanuele.chersoni@polyu.edu.hk}

\author[bloom]{Enrico Santus\corref{newaffiliation}}
\ead{esantus@bloomberg.net}

\author[uniud]{Giuseppe Serra}
\ead{giuseppe.serra@uniud.it}

\cortext[cor1]{Corresponding Author}
\cortext[newaffiliation]{The author was affiliated with Bayer Pharmaceuticals at the time of the experiments, and is currently affiliated with Bloomberg.}

\affiliation[uniud]{
    organization={AILAB Udine, Department of Mathematics, Computer Science and Physics, University of Udine}, 
    addressline={via delle Scienze 206},
    city={Udine},
    postcode={33100}, 
    state={Friuli-Venezia Giulia},
    country={Italy}
}

\affiliation[unina]{
    organization={Department of Biology, University of Naples Federico II},
    addressline={Corso Umberto I 40},
    city={80138},
    state={Campania},
    country={Italy}
}

\affiliation[polyu]{
    organization={Department of Chinese and Bilingual Studies (CBS), The Hong Kong Polytechnic University},
    city={Hung Hom},
    country={Hong Kong}
}

\affiliation[bayer]{
    organization={Bayer},
    state={New Jersey},
    country={USA}
}

\begin{abstract}
Adverse Drug Event (ADE) extraction is one of the core tasks in digital pharmacovigilance, especially when applied to informal texts. This task has been addressed by the Natural Language Processing community using large pre-trained language models, such as BERT.
Despite the great number of Transformer-based architectures used in the literature, it is unclear which of them has better performances and why.
Therefore, in this paper we perform an extensive evaluation and analysis of 19 Transformer-based models for ADE extraction on informal texts.
We compare the performance of all the considered models on two datasets with increasing levels of informality (forums posts and tweets).
We also combine the purely Transformer-based models with two commonly-used additional processing layers (CRF and LSTM), and analyze their effect on the models performance.
Furthermore, we use a well-established feature importance technique (SHAP) to correlate the performance of the models with a set of features that describe them: model category (\autoencoder, \autoregressive, \texttotext), pre-training domain, training from scratch, and model size in number of parameters.
At the end of our analyses, we identify a list of take-home messages that can be derived from the experimental data.
\end{abstract}




\end{frontmatter}



\section{Introduction}

In 2021, 50 new drugs were approved by the Food and Drug Administration (FDA) \cite{molecules27031075}, while 92 were recommended for marketing authorization by the European Medicines Agency (EMA) \cite{ema2021}.
The efficacy and safety of the newly-released medicines is verified through medical trials, which also have the purpose of identifying possible Adverse Drug Events (ADEs).
However, new collateral effects and adverse reactions might emerge once the medicinal is administered to a larger population of patients of different ages and medical conditions. 
To further safeguard the patients, Pharmacovigilance (PV) activities monitor all drugs after they entered the market, detecting and analyzing all ADEs reports.

Traditionally, the process of collecting ADEs relies on formal reporting methods (e.g., AERS, the Adverse Event Reporting System of the FDA), based on the communication between patients, healthcare providers, pharmaceutical companies, and local PV authorities.
ADEs can also be extracted (either manually of automatically) from formal medical documents, such as Electronic Health Records (EHR) (see \cite{feng2019using} for a recent overview).
However, studies show that such traditional reporting systems suffer from problems such as under-reporting: for example only 10\% of serious ADEs get registered in AERS \cite{wadman2005news}.

Recently, however, more and more social media users discuss their health status on forums and microblogging platforms, such as Facebook and Twitter. These posts include details regarding the users' physical and mental health, opinions on medications, and feedback on medical procedures.
This health-centric chatter generated on social media has the potential to become a new information channel, which works in parallel with the traditional reporting systems, to enhance the capabilities of digital PV systems \cite{sarker2015utilizing,karimi2015text}.
In fact, social media data could be used to collect the quasi-real-time feedback of the population during the roll-out of new drugs (e.g. COVID-19 vaccines during 2021) to promptly detect unexpected side-effects \cite{Portelli2022-lz}.

However, social media posts introduce several challenges due to the nature and structure of the texts, which differs a lot from formal EHRs.
In fact, posts, tweets, and messages in medical forums are usually highly informal, containing layman terms, typos, linguistic phenomena that could affect the meaning of the message. The same texts might also include specialized medical terms, drug names (both brand and generic ones), and mentions of medical conditions and procedures.

Given the complexity of the problem and the increasing need for automatic solutions, the topic of digital PV and ADE detection from social media texts has gained interest in the NLP community. A thematic workshop (Social Media Mining for Health -- SMM4H) has been organized since 2016 \cite{paul2016overview,sarker2017overview,weissenbacher2018overview,weissenbacher2019overview,klein-etal-2020-overview,magge-etal-2021-overview,smm4h2022overview}, to propose innovative solutions for ADE-related tasks on social media texts.
In this context, one of the core challenges is the ADE extraction task. It consists in tagging all spans of text representing an entity of interest inside a document, which in this case are Adverse Drug Events. For example, in the sentence 
\textit{``Fluoxetine and Quet combo \textit{zombified} me... ah, the meds merrygoround bipolar.''} we expect the system to extract the ADE \textit{zombified}.

This task is very complex for automatic systems due to the informal nature of the language and the presence of the aforementioned linguistic phenomena (e.g., humor, irony, speculations, negations), which can compromise the performance of current ADE extraction models \cite{scaboro2021nade, scaboro2022increasing}.

The proposed solutions were initially based on traditional machine learning, but then shifted to deep neural networks such as large language models. The latest proposed solutions employ a massive use of Transformer-based architectures \cite{vaswani2017attention}, especially the ones based on pre-trained models like \bert \cite{Devlin2019BERTPO}, and \bert variants trained on medical texts, such as \biobert \cite{lee2020biobert}, \xyendr \cite{tutubalina2017using}, and \bioroberta \cite{gururangan-etal-2020-dont}.
To further increase the final performance of the system, the models are frequently ensembled and often combined with additional processing modules such as BiLSTM \cite{graves2005framewise} and Conditional Random Field \cite{lafferty2001ConditionalRF} (CRF) \cite{papay2020dissecting}.

To the best of our knowledge, despite the great number of Transformer-based architectures used for ADE extraction in the literature, it is unclear which of these has the greatest benefits when used for this task. This raises the following questions:

\begin{itemize}
\item Which Transformer-based architectures (\autoencoder, \autoregressive, \texttotext) and variants work best for ADE extraction on informal texts?
\item What characteristics are shared by the best Transformer variants?
\item How do the different characteristics of the models (e.g., base architecture, the domain of the pre-training data) correlate with their performance?
\item What is the role played by the additional processing modules (e.g., BiLSTM and CRF) in the Transformers-based architectures?
\end{itemize}

To fill this gap, in this paper, we extensively compare 19 pre-trained Transformers-based models, ranging from the most traditional to the most recent ones, and from general-purpose ones to the ones specialized in the medical domain.
To be more thorough in our analysis, we decide to test the models on two different datasets, which represent different writing styles that can be encountered in online user-generated texts. The most informal writing style is represented by tweets, which are short, and contain slang numerous and non-standard orthography. We then chose forum posts as an example of longer social media texts, as they contain more complex sentences and detailed descriptions.
Using these two data sources with different textual styles allows us to better analyze the impact of the architectures of the models.
We also test the effect of additional processing modules (BiLSTM and CRF) in the architectures. Finally, we employ a well-known feature importance technique (\shapvals \cite{NIPS2017_7062}) to analyze the effect of the different model characteristics.

Our contribution can be summarized as follows:
\begin{itemize}
\item introduction of a unified framework to compare their predictions on the ADE extraction task, given the difference in output of \autoencoder and \autoregressive / \texttotext models; 
\item evaluation of the performance of the 19 pre-trained Transformer-based models on two well-known and stylistically different datasets;
\item analysis of the effect of commonly used additional processing modules for sequence labeling tasks (BiLSTM and CRF) and how they interact with the base models.
\end{itemize}
To guarantee the reproducibility of our experiments, we make publicly available\footnote{\url{https://github.com/AilabUdineGit/ade-detection-survey}} the source code used to perform the experiments and analysis presented in this paper.

The paper is organized as follows.
First, in the Related Work section, we present an overview of the methods commonly used for ADE extraction.
Next, in the Experimental Setting section, we describe the two datasets and the three model architectures used to address the task.
The paper continues with a description of the 19 Transformer variants that we are going to compare, the metrics used to evaluate them and a summary of the training details.
In the Results section, we present the evaluation of the models on the two datasets and an analysis to correlate the characteristics and performance of the models.
We conclude the paper with a final discussion of the results.

\section{Related Work}


In the literature, ADE extraction is usually framed as a Named Entity Recognition (NER) task, where the entity of interest is the ADE \cite{stanovsky2017recognizing}.
For this reason
the first solutions developed for this task were sequence labelling models based on traditional feature engineering and simple word embeddings, such as Word2Vec and GloVe \cite{Sarker2015PortableAT, Nikfarjam2015PharmacovigilanceFS}.
For example, Sarker et al. \cite{Sarker2015PortableAT} developed a probabilistic modelling method, which takes as input hand-crafted features extracted from the text, such as POS-tag, the presence of negations, the use of words belonging to specific vocabularies etc.

With the continuous progress of machine learning techniques and the introduction of the SMM4H shared task, methods based on neural networks became the most common choice for tackling the task.

With the advent of Transformers \cite{vaswani2017attention}, and the consequent development of large pre-trained language models (e.g., \bert \cite{Devlin2019BERTPO}, \gpttwo \cite{radford2019language}, \tfive \cite{https://doi.org/10.48550/arxiv.1910.10683}, \bart \cite{https://doi.org/10.48550/arxiv.1912.08777}, etc.), the ADE extraction community incorporated such models in new solutions, making them the building blocks of most of the top-performing systems. For example DeepADEMiner \cite{magge2021deepademiner} is a full deep learning pipeline to perform ADE extraction and normalization (i.e., mapping to medical ontologies) on tweets. It is comprised of a binary classifier based on \roberta, an ADE extractor based on \distilbert and an ADE normalizer based on \bert.

We can easily visualize how the proposed models became more Transformer-oriented looking at the architectures proposed to solve the SMM4H ADE extraction task, which was first introduced in 2019. Each year the top-2 models have always been Transformer-based (see Table \ref{tab:related_smm}), however the overall presence of Transformer-based models in the shared task has changed greatly.

\begin{table}[!htbp]
\resizebox{\linewidth}{!}{
\centering \scriptsize
\begin{tabular}{l c c p{.4\linewidth}}
\hline
\textbf{Ref.} & \textbf{Model} & \textbf{\specialcell{Additional\\ Resources}} & \textbf{Notes} \\ \hline
\cite{miftahutdinov-etal-2019-kfu} & \specialcell{\biobert\\+ CRF}  & \specialcell{External\\dictionaries\\CADEC} & Ensemble of 10 models to improve robustness \\
\cite{ge-etal-2019-detecting}      & \specialcell{Character-level CNN\\+ Word-level BiLSTM\\+ Multi-head self-attention\\+ CRF} & \specialcell{\\Word2Vec emb.\\ELMo emb.\\POS tagging\\Sentiment lexicon\\SIDER lexicon} & Use of several additional features and embeddings \\ \hline
\cite{miftahutdinov-etal-2020-kfu} & \xyendr  & \specialcell{External\\dictionaries\\CADEC} & -- \\
\cite{gattepaille-2020-far}        & \bert   & -- & Training only on tweets with at least one ADE mention, padding/truncation to 50 tokens \\ \hline
\cite{dima-etal-2021-transformer}  & \biobert  & \specialcell{Data\\augmentation} & Multi-task learning (binary classification + extraction + normalization), the first 11 layers of \biobert are frozen, three to five binary classifiers are ensembled to improve robustness \\
\cite{yaseen-langer-2021-neural}   & \specialcell{BiLSTM+CRF\\+\roberta emb.} & \specialcell{FastText emb.\\Byte-Pair emb.\\POS tagging} & Ensemble of 3 models to improve robustness \\ \hline
\cite{liu-etal-2022-pingantech}    & \specialcell{W2NER\\(\bert+LSTM+CNN)\\} & -- & Character and location features \\
\cite{guellil-etal-2022-edinburgh} & DeepADEMiner (\roberta) & Flair emb. & -- \\ \hline
\end{tabular}
}
\caption{Details on the top-2 models in the SMM4H workshops (years 2019-2022).}
\label{tab:related_smm}
\end{table}

In 2019, 50\% of the proposed models (5 out of 10) were based on traditional deep learning models. For example, the second-best architecture \cite{ge-etal-2019-detecting} was based on Convolutional Neural Networks (CNNs), BiLSTMs, CRF and Multi-head self-attention, employing features such as part-of-speech tagging, ELMo embeddings \cite{peters-etal-2018-deep}, and Word2Vec embeddings \cite{mikolov2013efficient}. Sarabadani \cite{sarabadani-2019-detection} also used LSTMs and CNNs, combined with ELMo embeddings and three specialized lexicon sets, while Lopez et al. \cite{lopez-ubeda-etal-2019-using-machine} used a CRF with GloVe embeddings \cite{pennington2014glove}. The other half of the proposed models were all based on the recently-introduced BERT and its variants, including the best architecture for 2019 \cite{miftahutdinov-etal-2019-kfu}, which employed an ensemble of BioBERTs with a CRF module.

In the 2020 SMM4H edition, 66\% of the proposed models (4 out of 6) were based on Transformers, and the three best architectures were based on BERT \cite{gattepaille-2020-far} or multilingual \autoencoder models such as \xyendr \cite{miftahutdinov-etal-2020-kfu} (pre-trained on an English collection of consumer comments on drug administration) and \roberta \cite{kalyan-sangeetha-2020-want}.

Finally, in 2021 and 2022, 100\% of the teams who provided system descriptions used Transformer-based models. The top architectures in SMM4H 2022 combined them with additional features, such as character and location features \cite{liu-etal-2022-pingantech}, or Flair embeddings \cite{guellil-etal-2022-edinburgh}. The third-best architecture used an ensemble of 5 BERT-large models to increase the system's robustness \cite{candidato-etal-2022-air}, while the fourth team \cite{uludogan-yirmibesoglu-2022-boun} was the first to report using \gpttwo, a \texttotext model, during these shared tasks.

%
In the last years, \texttotext approaches based on Transformers have been proposed \cite{https://doi.org/10.48550/arxiv.2109.05815} to solve the ADE extraction task with promising results on several datasets, including generalizability across text genre and some zero-shot cross-language transfer capabilities.

\medskip

Since Transformer-based models showed great results in medical-domain NLP, Wang et al. \cite{Wang2021} compiled an extensive survey of their use in the biomedical domain, including an overview of tasks and architectures. However, this work does not include a practical performance evaluation of the models and, in particular, it does not cover the topic of ADE extraction on social media.

Instead, in this paper, we perform an extensive comparison of Transformers-based architectures for ADE extraction on social media texts. To perform a more complete analysis, we take into consideration the three main categories of Transformer-based models: \autoencoder, \autoregressive and \texttotext models.

\section{Material and methods}

With the aim of performing a systematic analysis of Transformer-based architectures in the context of ADE extraction, in this section we report the details of the experimental setting put in place.
We introduce the 19 Transformer variants used for the task of ADE extraction and the two benchmark datasets with different grades of informality and different textual styles.
We also illustrate how the two additional processing modules (LSTM and CRF) are incorporated in the experiments.
Finally, we describe the methodology used to perform the feature importance analysis using \shapvals, to correlate the models features and their performance.

\subsection{Datasets}

Due to the strong interest of the research community on the task of ADE extraction, over the years several corpora containing informal texts have been released \cite{Alvaro2017TwiMedTA,Sarker2015PortableAT,Karimi2015CadecAC,klein-etal-2020-overview,webradr}.
Among all these datasets, we selected the two most widely used ones: CADEC \cite{Karimi2015CadecAC} and SMM4H \cite{klein-etal-2020-overview}. They are the largest and most updated datasets for ADE extraction on social media texts, fully annotated for the presence of ADEs and widely used by the community. These datasets also present two different textual typologies, which allows us to perform a comparative analysis of different kinds of social media data.

Indeed, CADEC is composed of long and structured messages from medical forum reports, while SMM4H contains highly informal texts coming from Twitter.

To verify the difference in textual style, we extract some statistics from the texts of the two datasets and report them in Table \ref{tab:textstats}: the count of syllables, lexicon (how many different word types are being used), sentences, characters, and the number of difficult words per samples. "Difficult words" refers to the number of polysyllabic words with Syllable Count $> 2$ that are not included in the list of words of common usage in English. We calculate the same metrics for the full texts of the samples, and the ADEs.
Table \ref{tab:textstats} shows that the CADEC dataset contains significantly longer texts and more complex words (14 versus 4 Difficult Words per sample). The ADE mentions in CADEC are also longer (4.06 syllables versus 1.32 syllable on SMM4H), and there are more ADE mentions per sample (5.40 versus 1.62).

\begin{table}[!htbp]
\small
\centering
\begin{tabular}{cl r@{ $\pm$ }l r@{ $\pm$ }l}
\hline
\multicolumn{2}{c}{\textbf{Metric}} &
\multicolumn{2}{c}{\textbf{CADEC}} &
\multicolumn{2}{c}{\textbf{SMM4H}} \\ \hline
\multirow{6}{*}{\rotatebox[origin=c]{90}{\textbf{Full text}}}
& Syllable Count  & 116 &  2.7 & 25 &  8.6 \\
& Lexicon Count   &  83 &  1.9 & 17 &  6.1 \\ 
& Sentence Count  &   5 &  0.1 &  2 &  0.9 \\
& Character Count & 461 & 10.5 & 86 & 28.6 \\ 
& Difficult Words &  14 &  0.3 &  4 &  2.1 \\ 
\cline{2-6}
& Number of ADEs  &  5.40 & 4.5 & 1.62 & 0.7 \\
\hline
\multirow{4}{*}{\rotatebox[origin=c]{90}{\textbf{ADE}}}
& Syllable Count  &  4.06 & 2.7 & 1.32 & 1.8 \\
& Lexicon Count   &  2.62 & 1.9 & 0.89 & 1.3 \\ 
& Character Count & 14.07 & 8.5 & 4.78 & 6.1 \\ 
& Difficult Words &  0.89 & 0.8 & 0.29 & 2.1 \\ 
\hline
\end{tabular}
\caption{Average textual metrics for the two datasets, computed with the \textsc{textstat}\cite{textstat} Python library. The readability metrics are calculated on the full text of the documents (first 5 rows) and on the ADEs only.}
\label{tab:textstats}
\end{table}

\paragraph*{CADEC} The dataset contains 1250 posts from the health-related forum ``AskaPatient''\footnote{\url{https://www.askapatient.com/}}, where the users report their ADEs. A total of 1107 posts contain at least one ADE (positive samples), while the remaining 143 do not contain any ADE mention (negative samples). The language used in this forum posts is generally informal, frequently deviating from standard English. For the training and evaluation we use the splits made publicly available by \cite{dai2020EffectiveTransition}.

\paragraph*{SMM4H} The dataset is composed of English-language tweets containing a drug name and possibly an ADE. We use the annotated data of the ADE Extraction Task of the SMM4H 2020 shared task, which contains {1862} tweets, {1080} of which are positive for the presence of ADEs while the remaining {782} are negative. Similarly to previous works \cite{portelli-etal-2021-bert, SAKHOVSKIY2022104182}, we only use the annotated samples provided by the shared task (training and validation set) and not the blind test set for our analyses. The evaluation on the blind test set is available through the CodaLab platform\footnote{\url{https://competitions.codalab.org/competitions/23705\#results}}, however CodaLab allows for a limited number of test runs. Since our work entails a large number of experiments with multiple base models, combinations with extra modules, and multiple seeds, this would create a large amount of traffic on the platform, long queues to get the results, and could reach the run limit. Furthermore, using the blind test set would not allow us to compute additional metrics or perform in-depth error analyses on the models predictions. Therefore, we only use the annotated train and validation sets. The available samples are partitioned into new train, validation, and test sets\footnote{Splits available at \url{https://github.com/AilabUdineGit/ADE}}. Each set contains the same proportion of texts with and without ADEs.

\paragraph*{Data Preprocessing}

In both datasets, the presence of an ADE is annotated at the character level with a list of $(start, end)$ annotations indicating that the ADE entity begins at the character $start$ and spans until the character $end$ (excluded). Following the previous literature, we converted the annotations using the Begin-Inside-Outside (\textit{BIO}) annotation scheme for the tokens (words that compose a text), where \textit{B} marks the beginning of an entity, \textit{I} the following tokens belonging to the entity and \textit{O} marks the fact that the token does not belong to an ADE.

Some specialized preprocessing steps were necessary due to the different tagging procedures used in the two datasets.
The annotation scheme of CADEC allows for the presence of discontinuous and/or overlapping entities, meaning that the ADE might be composed by non-consecutive pieces of text (e.g., ``I felt an \uline{intense}, even if expected, \uline{nausea}'' $\rightarrow$ ``intense nausea'') or the same piece of text could belong to two different ADEs (e.g., ``I felt intense \uline{pain in the hip} and \uline{right foot}''  $\rightarrow$ ``pain in the hip'', ``pain in the right foot'').
The customary solution is to disambiguate the annotations, merging overlapping ADEs and separating discontinuous mentions, which constitute about $10\%$ of mentions in CADEC \cite{dai2020EffectiveTransition}\footnote{Some past works proposed alternative NER-based approaches to deal with these kinds of annotations without disambiguation \cite{dai2018recognizing,dai2020EffectiveTransition}.}.
Both datasets were preprocessed to disambiguate overlapping and discontinuous annotations.

\subsection{Model Architectures}

The analyzed models belong to three macro-categories, \autoencoder, \autoregressive and \texttotext, which need different architectural choices to address the task.

\autoencoders are the most commonly used model for the task of ADE extraction, while \autoregressive and \texttotext models, which produce textual outputs, have only recently been tested on ADE extraction \cite{https://doi.org/10.48550/arxiv.2109.05815}.

\subsubsection{\autoencoder models}

The first category of models we consider are the \autoencoders. With the term \autoencoder model we mean an architecture that is composed of a stack of Transformer encoders. This stack produces as output a series of embeddings. At the top of this architecture, other layers can be added to solve a particular task. In this case, we add a Linear Layer to project the sequence of embeddings to a probability distribution over the output labels (BIO labels).
Finally, the actual output is calculated for each input word (token). More precisely, given a sentence $s=t_1,\dots,t_n$, where $n$ is the sentence length and $t_i$ is the $i$-th token, we perform token classification to extract ADEs in the following way:
\begin{align*}
    h & = M(s) &h \in \mathbb{R}^{n\times 768} \\
    a & = Wh+b  &W \in \mathbb{R}^{768 \times 3} \\
    y_i & = \frac{e^{a_i}}{\sum_i^n e^{a_i}} \\
    \ell_i & = \arg\max(y_i)
\end{align*}

Where $\ell_i$ is the predicted label for the $i$-th token $t_i$ and $M$ is the \autoencoder model.

\begin{figure}[!hbtp]
    \centering
    \includegraphics[width=0.8\linewidth]{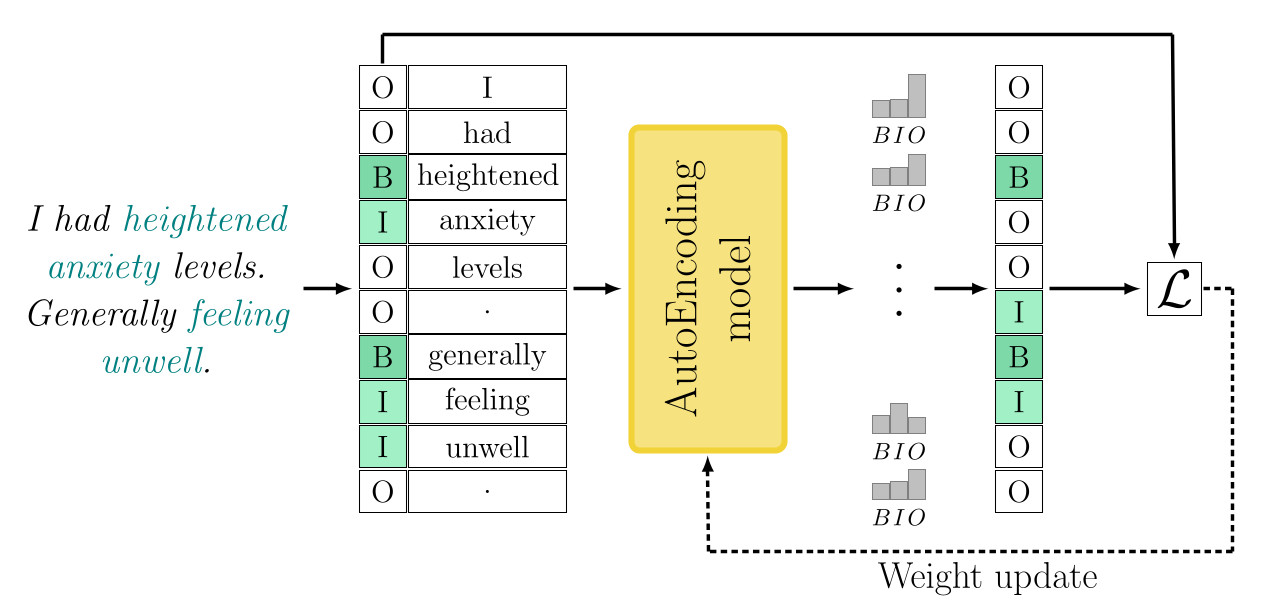}
    \caption{The ADE extraction pipeline for \autoencoders.}
    \label{fig:base_architecture_a}
\end{figure}

This base architecture and the training procedure are shown in Figure~\ref{fig:base_architecture_a}.

Following the literature on this task, we experiment and combine the \autoencoder models with two additional processing layers: Conditional Random Fields (CRF)  \cite{lafferty2001ConditionalRF} and bidirectional LSTMs (BiLSTM) \cite{graves2005framewise}.

The {\crf{\autoencoder}}
architecture combines the Transformer model with a CRF classifier. The BIO probability distribution generated by the Transformer model becomes the input of a CRF module, which produces another sequence of subword-level BIO labels. This step aims at denoising the sub-word output labels produced by the previous component.

The {\lstm{\autoencoder}}
architecture combines the Transformer model with a BiLSTM. The embeddings generated by the Transformer model become the input of a one-layer BiLSTM that produces new embeddings of the same size. These new representations are then passed to a Linear Layer + Softmax, turning them into a probability distribution over the BIO labels.

\subsubsection{\autoregressive and \texttotext models}

\autoregressive and \texttotext models work similarly. Both kinds of models take a text as input and return a text as output.
However, \autoregressive models are composed of a stack of Transformers decoders, while \texttotext models use the entire decoder-decoder architecture of the original Transformer \cite{vaswani2017attention}.

\begin{figure}[!hbtp]
    \centering
    \includegraphics[width=0.7\linewidth]{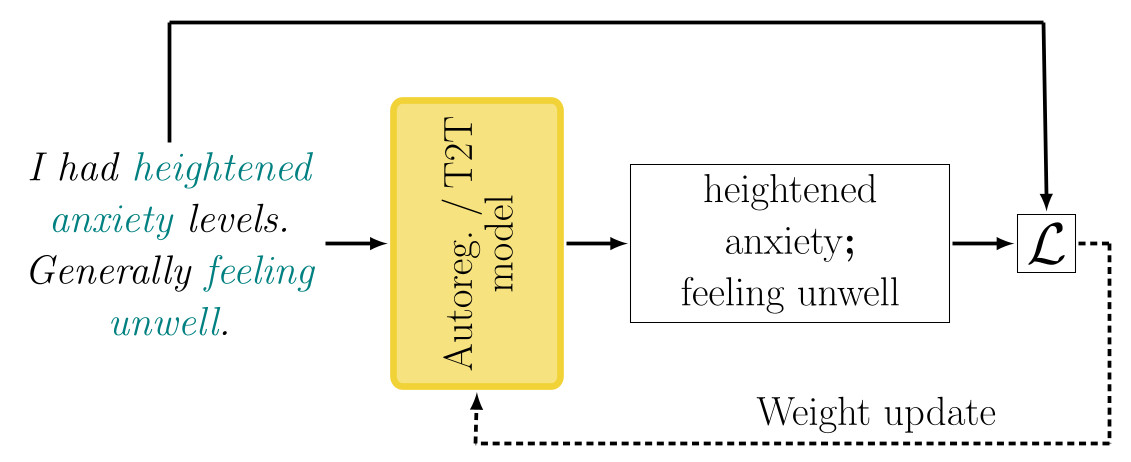}
    \caption{The ADE extraction pipeline for \autoregressive and \texttotext models.}
    \label{fig:base_architecture_b}
\end{figure}

We train the models to produce as output the list of the ADEs present in the input text, separated by semicolons. This architecture and the training steps are shown in Figure \ref{fig:base_architecture_b}.

\begin{figure}
    \centering
    \includegraphics[width=0.35\linewidth]{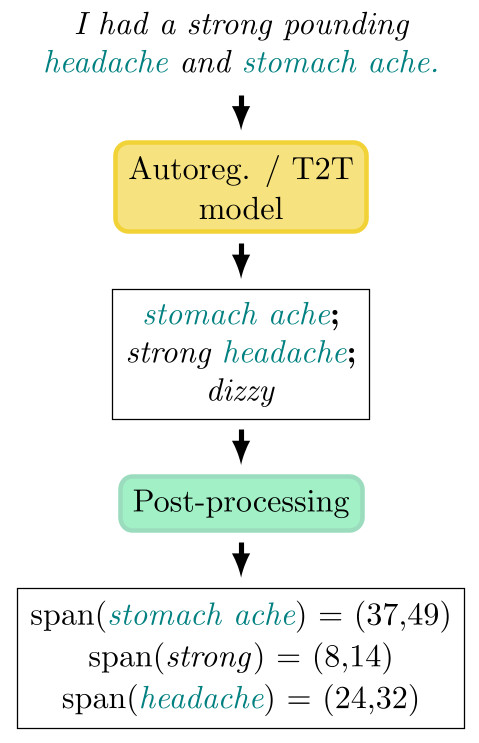}
    \caption{Example of the post-processing procedure used to map the string output of the \autoregressive and \texttotext models to a list of ADE entities contained in the input text.}
    \label{fig:postproc}
\end{figure}

To evaluate the performance of the model, it is necessary to map its output back to the original text, however there is no guarantee that the strings produced by the model are exact substrings of the original text. Therefore, a simple postprocessing procedure is used to map the list of output ADEs to the input text. Let us consider the example in Figure \ref{fig:postproc}.
Each item in the semicolon-separated output can contain more than one word.
If the item is a perfect sub-string of the input text, we consider it as single prediction.
This is the case for ``stomach ache'', which becomes span 1 after postprocessing.
If the item is not a perfect sub-string if the input, we split it into shorter substrings that belong to the text and consider them as separate prediction.
For example, the item ``strong headache'' gives origin to two predictions: spans 2 and 3.
If part of the item cannot be found in the original text, such as ``dizzy'' in our examples, that part is completely discarded and does not generate a prediction.

\subsection{Transformer Variants}
\label{sec:models}

In this section, we briefly present the 19 Transformer-based model variants chosen for this survey, illustrating their main features. We start with all the models trained on general-domain texts only and then move to the variants that use in-domain knowledge, either medical or coming from social media data.\footnote{There is a great number of \autoencoder and \autoregressive pre-trained or fine-tuned on in-domain datasets. We have selected the most relevant and diverse ones to include in the analysis. Other models present in the literature would have been an interesting addition (e.g., \medgpttwo \cite{https://doi.org/10.48550/arxiv.2107.03134}, a \gpttwo model fine-tuned on EHRs), but could not be included due to lack of public code and model checkpoints.}

Notice that three of the in-domain variants were pre-trained from scratch (\scibert, \pubmedbert, and \bertweet), meaning that they have a unique vocabulary tailored to their pre-training corpus and include specific embeddings for in-domain words.

Table \ref{tab:info_models} is a summary of the information about the version of all Transformer-based models used. The upper part of the Table lists general-domain variants (Section \ref{sec:variants_general}), while the lower part lists variants with in-domain knowledge (Section \ref{sec:variants_domain}). The first column reports the model's category (\autoencoder, \autoregressive or \texttotext). The column ``From Scratch'' marks which models were trained from scratch, as opposed to the ones which were initialized with another model's weights (e.g., \roberta was trained from scratch while \bioroberta was initialized with \roberta's weights and therefore shares part of its knowledge). The three columns under the name ``Pre-training Domain'' record the kind of documents which the models were pre-trained on: General domain knowledge (e.g., Wikipedia or BookCorpus), Medical domain (e.g., PubMed full-texts or health records), and Social domain (e.g., tweets or forum posts). For example, \roberta was pre-trained on General-domain documents only, while \bioroberta has both General-domain knowledge (derived from \roberta's pre-training) and Medical-domain knowledge (derived from its own additional pre-training). Finally, the Table reports the model's size in millions of parameters.

\begin{table*}[!hbtp]
\centering
\resizebox{\linewidth}{!}{%

\begin{tabular}{ r | c | c | ccc | c }
\hline
\multicolumn{1}{c}{} &
\multicolumn{1}{c}{} &
\multicolumn{1}{c}{\textbf{From}} &
\multicolumn{3}{c}{\textbf{Pre-training Domain}} &
\multicolumn{1}{c}{\textbf{Model}} \\

\multicolumn{1}{c}{\textbf{Model Name}} &
\multicolumn{1}{c}{\textbf{Category}} &
\multicolumn{1}{c}{\textbf{Scratch}} &
\textbf{General} & \textbf{Medical} & \multicolumn{1}{c}{\textbf{Social}} &
\multicolumn{1}{c}{\textbf{Size}} \\ \hline

\bert            & \autoencoder    & $\times$ & $\times$ &          &          & 109M \\
\distilbert      & \autoencoder    &          & $\times$ &          &          &  66M \\
\spanbert        & \autoencoder    & $\times$ & $\times$ &          &          & 108M \\
\roberta         & \autoencoder    & $\times$ & $\times$ &          &          & 124M \\
\electra         & \autoencoder    & $\times$ & $\times$ &          &          & 109M \\
\xlnet           & \autoregressive & $\times$ & $\times$ &          &          & 118M \\
\gpttwo          & \autoregressive & $\times$ & $\times$ &          &          & 124M \\
\tfive           & \texttotext     & $\times$ & $\times$ &          &          & 223M \\
\pegasus         & \texttotext     & $\times$ & $\times$ &          &          & 570M \\
\bart            & \texttotext     & $\times$ & $\times$ &          &          & 139M \\ \hline
\bertweet        & \autoencoder    & $\times$ &          &          & $\times$ & 354M \\
\biobert         & \autoencoder    &          & $\times$ & $\times$ &          & 109M \\
\bioclinicalbert & \autoencoder    &          & $\times$ & $\times$ &          & 108M \\
\scibert         & \autoencoder    & $\times$ &          & $\times$ &          & 109M \\
\pubmedbert      & \autoencoder    & $\times$ &          & $\times$ &          & 108M \\
\xyendr          & \autoencoder    &          &          & $\times$ & $\times$ & 177M \\
\bioelectra      & \autoencoder    & $\times$ &          & $\times$ &          & 109M \\
\bioroberta      & \autoencoder    &          & $\times$ & $\times$ &          & 124M \\
\scifive         & \texttotext     &          & $\times$ & $\times$ &          & 223M \\ \hline
\end{tabular}

}
\caption{Information about the version of all the Transformer-based models used and their pre-training.}
\label{tab:info_models}
\end{table*}

\subsubsection{General-domain Variants}
\label{sec:variants_general}

\modeltitle{\bert} \cite{Devlin2019BERTPO}{, \autoencoder. Standard model, pre-trained on general-domain texts (Wikipedia and BookCorpus) with two objectives: Masked Language Modeling (MLM) and Next Sentence Prediction (NPS). In MLM, a token in the input sentence is replaced with the MASK token and the goal of the model is to identify the original one. In NSP the model classifies the second input sentence as related or not to the first one. As mentioned in the related work, \bert achieved state-of-the-art results in several NLP tasks and worked as the foundation of many other pre-trained models.}

\modeltitle{\distilbert} \cite{Sanh2019DistilBERTAD}, \autoencoder. It is a distilled version of the original \bert model. A student network, with half the number of layers of \bert, is initialized with the weights of its \bert teacher, taking one layer out of two. The student model is then trained to replicate the output distribution of the teacher using three losses: Masked Language Modeling (MLM), distillation loss (CE), and cosine embedding loss (COS).

\modeltitle{\spanbert} \cite{Joshi2019SpanBERTIP}, \autoencoder. A version of \bert that introduces an additional loss called Span Boundary Objective (SBO), alongside the traditional MLM loss used for \bert.\\
Let us consider a sentence $S = [w_1, w_2, \dots, w_k]$ and its sub-string $S_{m:n} = [w_m, \dots, w_{n}]$. $w_{m-1}$ and $w_{n+1}$ are the boundaries of $S_{m:n}$ (the words immediately preceding and following it). We \textit{mask} $S$ by replacing all the words in $S_{m:n}$ with the \texttt{[MASK]} token. \spanbert reads the masked version of $S$ and returns an embedding for each word. The MLM loss measures if it is possible to reconstruct each original word $w_i \in S_{m:n}$ from the corresponding embedding. The SBO loss measures if it is possible to reconstruct each $w_i \in S_{m:n}$ using the embeddings of the boundary words $w_{m-1}$ and $w_{n+1}$. This kind of pre-training procedure makes its embeddings more appropriate for NER-like tasks.

\modeltitle{\roberta} \cite{Liu2019RoBERTaAR}, \autoencoder. {Starting from the assumption that \bert was under-trained, \roberta was developed changing some aspects of \bert's pre-training phase.
It dynamically changes the masking pattern: instead of using a static masking strategy, each training sample was duplicated 10 times, masking each sequence in 10 different ways. Additionally, \roberta is trained without the NSP objective, for more steps, with more data, bigger batches, and on longer text sequences.
This model achieved state-of-the-art performances surpassing \bert on many general-domain NLP tasks.}

\modeltitle{\electra} {\cite{clark2020electra}, \autoencoder. It is a pre-trained model where the MLM objective is replaced with the Replaced Token Detection task, in which the model learns to distinguish real input tokens from synthetically generated replacements. The network is trained as a discriminator that predicts for every token whether is original or a replacement. The role of the generator is usually covered by a small MLM model. In this way, the model gains knowledge from all input tokens instead of just the small masked-out subset. This approach keeps the performances close to the ones of \bert, while lowering the computational costs of training the model.}

\modeltitle{\xlnet} {\cite{yang2020xlnet}, \autoregressive. It is a pre-trained model that tries to leverage the best of both \autoregressive and \autoencoder language modeling. Instead of using a fixed forward or backward factorization order, it maximizes the expected log-likelihood of a sequence with respect to all possible permutations. This objective is called Permutation Language Modeling. A difference with \bert is that this model does not rely on data corruption (e.g. token masking).}

\modeltitle{\gpttwo} {\cite{radford2019language}, \autoregressive. It is a stack of transformer-decoders pre-trained with the simple objective of Next Word Prediction. As mentioned in the related work, it achieved state-of-the-art results on several text completion benchmarks.}

\modeltitle{\tfive} {\cite{https://doi.org/10.48550/arxiv.1910.10683}, \texttotext. It is an encoder-decoder model pre-trained on a multi-task mixture of unsupervised and supervised tasks. The are some small differences between T5 and the classical Transformer encoder-decoder. An example is the use of a normalization layer after each layer in both the encoder and decoder. The Span-based language masking objective is used during the pre-training phase, masking some randomly selected words in the input sentence and generating those words separated by the masking token \textit{$<$M$>$}. The model has been trained using the C4 corpus \cite{dodge2021documenting}.}

\modeltitle{\pegasus} {\cite{https://doi.org/10.48550/arxiv.1912.08777}, \texttotext. It is an encoder-decoder model pre-trained using a self-supervised objective (gap-sentence-objective) and created originally to improve the fine-tuning performance on abstractive summarization. In our case, the model is used in a text-generation setting and not specifically for a summarization task.}

\modeltitle{\bart} {\cite{https://doi.org/10.48550/arxiv.1910.13461}, \texttotext. It is a model pre-trained using two strategies: corrupting text by shuffling the original order of sentences and masking spans of text by replacing them with a mask token. It matches the performance of \roberta on several NLP benchmarks that require text comprehension, and is effective in text-generation tasks.}

\subsubsection{Domain-specific Variants}
\label{sec:variants_domain}

\modeltitle{\bertweet} {\cite{nguyen2020bertweet}, \autoencoder. The model is trained \textit{from scratch} using the same pre-training procedure of \roberta and a dataset containing 873M tweets. Some of them belong to the general Twitter Stream grabbed by the Archive Team\footnote{\url{https://archive.org/details/twitterstream}}, while others are related to the COVID-19 pandemic. We use the large version that allows us to input up to 512 tokens, analyzing the longer CADEC texts.}

\modeltitle{\biobert} \cite{lee2020biobert}, \autoencoder. The model was pre-trained on PubMed abstracts starting from a \bert checkpoint.
The authors of \biobert provide different versions of the model, pre-trained on different corpora. We selected the version which seemed to have the greatest advantage on this task, according to the results by \cite{lee2020biobert}. We chose \biobert v1.1 (+PubMed), which outperformed other \biobert v1.0 versions (including the ones trained on full texts) in NER tasks involving Diseases and Drugs. Preliminary experiments against \biobert v.1.0 (+PubMed+PMC) confirmed this behavior on the datasets used in this work \cite{portelli-etal-2021-bert}.

\modeltitle{\bioclinicalbert} \cite{alsentzer2019publicly}, \autoencoder. It was pre-trained on clinical texts from the MIMIC-III database \cite{mimiciii}, starting from a \biobert checkpoint.

\modeltitle{\scibert} \cite{beltagy2019scibert}, \autoencoder. It was pre-trained \textit{from scratch} on papers retrieved from Semantic Scholar \cite{ammar-etal-2018-construction} (82\% of them belonging to the medical domain).

\modeltitle{\pubmedbert} \cite{pubmedbert}, \autoencoder. It was pre-trained \textit{from scratch} on PubMed abstracts and full-text articles from PubMed Central\footnote{\url{https://www.ncbi.nlm.nih.gov/pmc/}}.
The vocabulary of \pubmedbert contains more in-domain medical words than any other model under consideration (as reported in their paper).

\modeltitle{\xyendr} \cite{tutubalina2017using}, \autoencoder. The model was pre-trained on an English corpus of health-related comments \cite{tutubalina2017using} starting from a BERT base multilingual cased checkpoint.

\modeltitle{\bioroberta} {\cite{gururangan-etal-2020-dont}, \autoencoder. It was pre-trained from a \roberta-base checkpoint on biomedical full-text papers from S2ORC \cite{lo2020s2orc}.}

\modeltitle{\bioelectra} {\cite{Kanakarajan2021BioELECTRAPretrainedBT}, \autoencoder. It was pre-trained \textit{from scratch} on clinical texts from PubMed abstracts using the same architecture as \electra.}

\modeltitle{\scifive} {\cite{https://doi.org/10.48550/arxiv.2106.03598}, \texttotext. It is a domain-specific \tfive model pre-trained on a large biomedical corpus of PubMed Abstracts and PMC articles, starting from a \tfive checkpoint.}

\subsection{Metrics}
\label{sec:metrics}

Since the problem is framed as either multi-class token classification (BIO labels) or text generation, which eventually outputs a set of predicted entities, we use the standard evaluation metrics used by the ADE extraction community, which are entity-level relaxed F1 score, Precision and Recall. The following describes in detail how the metrics are calculated.

\medskip

\begin{figure}[!hbtp]
\centering
\includegraphics[width=0.7\linewidth]{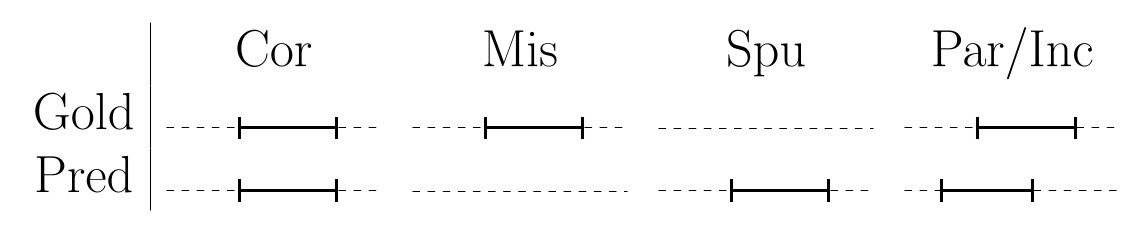}
\caption{Visual representation of the intermediate metrics used to calculate Precision, Recall, and F1 score. The schema compares the presence of real annotations (Gold) and the predictions of the model (Pred).}
\label{tab:metrics}
\end{figure}

Given a set of gold (ground-truth) entities and a set of predicted entities, we can calculate the following values (see Figure \ref{tab:metrics}): \textit{Correct} ($Cor$), the number of entities which perfectly correspond to the gold ones; \textit{Missing} ($Mis$) all gold entity not present in the predictions; \textit{Spurious} ($Spu$) number of excess predicted entities; \textit{Partial}/\textit{Incorrect} ($Par$/$Inc$) the number of predicted entities which partially overlap a gold entity. In practice, one of the last two values is set to 0: $Par=0$ if we want to consider partially overlapping entities as incorrect, while $Inc=0$ if consider them as correct.

Starting from these values, we define the main evaluation metrics used for this task, which are the Strict and Relaxed versions of the F1, Precision, and Recall scores \cite{Semeval2013}, calculated at the entity level \cite{chinchor-sundheim-1993-muc}.

The Relaxed versions of the metrics, which allow for partial overlaps, are defined as follows:

\begin{align*}
Recall &= \frac{Cor + (Par\times 0.5)}{Cor+Par+Inc+Mis}\\\\
Precision &= \frac{Cor + (Par\times 0.5)}{Cor+Par+Inc+Spu}\\\\
F1 &= \frac{2\times Precision \times Recall}{Precision + Recall}
\end{align*}\\

The Strict Recall, Precision, and F1 score are calculated using the same formulas above, setting $Par=0$.

Relaxed and Strict metrics are highly correlated and follow the same trends. In this work, when commenting results we will always refer to the Relaxed metrics to keep the discussion concise and avoid repetitions. The Strict metrics are reported in \ref{app:full_metrics}.

\subsection{Feature Importance Analysis}

Our objective is to analyze how some high-level features of the models correlate with their final performance (F1 score). For this reason, we characterize each model with the following six features:

\begin{enumerate}
\item Model Category: \autoencoder (0), \autoregressive (1), \texttotext (2);
\item Pre-training domain - General data: Yes, No;
\item Pre-training domain - Medical data: Yes, No;
\item Pre-training domain - Social data: Yes, No;
\item Pre-training from scratch: Yes, No;
\item Model Size (number of parameters): less than 100M (0), 100M--130M (1), over 130M (2).
\end{enumerate}

The values of these features of all the models can be derived from Table \ref{tab:info_models}. We encode Model Category using label encoding, as opposed to one-hot encoding.
We prefer label encoding to one-hot encoding because it helps to better highlight and analyze the effect of the three values of this feature. Using one-hot encoding would split its contribution among three separate features and make it harder to see their interaction with the chosen technique.
We verified that the ordering chosen to encode the features does not impact the results by permuting the values used to encode the Model Category and comparing all the results. 

To explain the performance of the models starting from this set of high-level features, we employ \shapvals \cite{NIPS2017_7062}, a widely-used model interpretability technique.
This technique assigns a positive or negative \textit{contribution} to each value of all the input features, representing the impact that they have on the model's output.
 
To calculate the \shapvals, we need to create a model that takes as input the six high-level features previously listed and outputs the F1 score of the Transformer variant.
We choose a Random Forest model, as it is well-suited to work on low-dimensional data and it is also often used to perform these kinds of analyses.
To generate a high number of input data to fit the Random Forest model, we use the results of all 30 runs of the previous experiments to calculate the performance of each Transformer-based variant.
Therefore, we obtain 570 (30 $\times$ 19) samples containing the six high-level features and the F1 score of the models.

\begin{figure}
    \centering
    \includegraphics[width=0.73\linewidth]{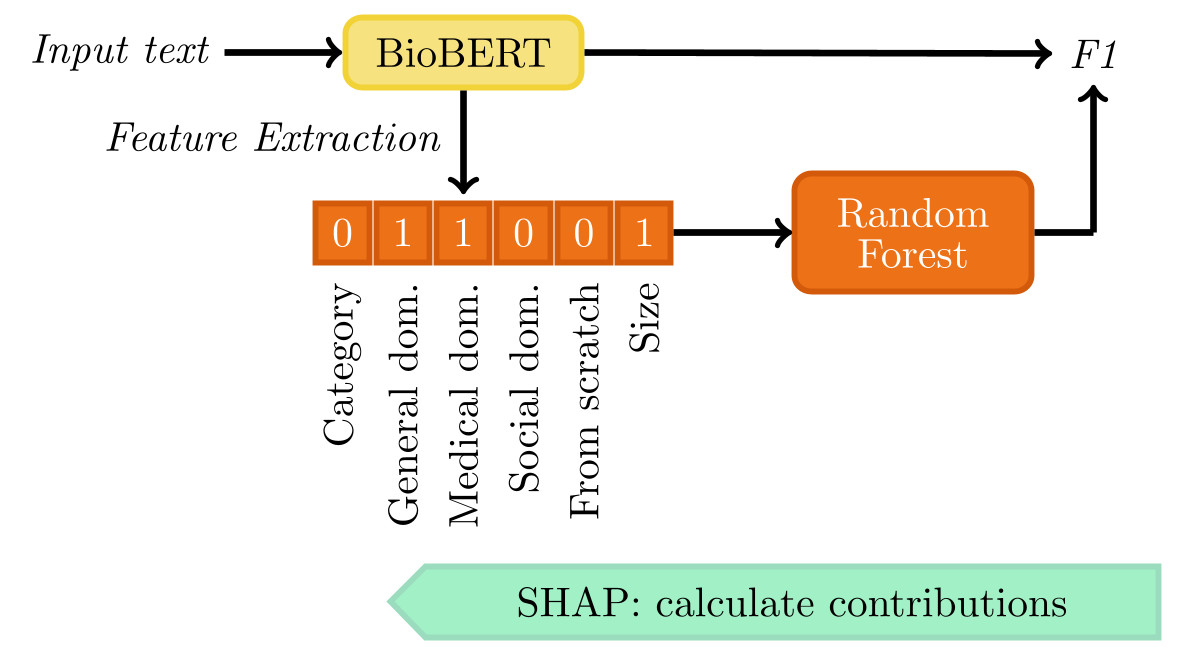}
    \caption{Process used to generate the \shapvals for the Transformer-based models.}
    \label{fig:shapval_procedure}
\end{figure}

Figure \ref{fig:shapval_procedure} summarizes the process used to generate the \shapvals.

After fitting the Random Forest, we can use the same set of data to calculate the attributions for each input feature and their values.

\subsection{Training details}

For all models, we performed hyperparameter tuning via grid-search. The models were evaluated on the training set and the best hyperparameters were chosen based on the highest relaxed F1 score.

We tested the following parameters:
\begin{itemize}
\item learning rate: $[5e{-5}, 5e{-6},1e{-3}, 1e{-4},1e{-5}]$
\item dropout rate: from $0.15$ to $0.30$, increments of $0.05$
\item batch size: $[8,16,32]$ for SMM4H, $[4,8]$ for CADEC
\item training epochs: $1$ to $15$
\end{itemize}

The best hyperparameters selected for all the models are reported in \ref{app:hyperparam}.

The input sequence length was fixed to $512$ for the CADEC dataset and $64$ for the SMM4H dataset.

\autoencoder and \texttotext models were allowed to generate sequences with a maximum length of $40$ tokens for CADEC and $20$ tokens for SMM4H, based on the expected output sequence length in the training set.

During the final evaluation, all the models were tested on the test set, after being trained with the best hyperparameters on the concatenation of the training set and the validation set. The evaluation was repeated thirty times with different random seeds. We report the average of the results over the thirty runs.

Both training and testing have been performed using an Nvidia GeForce 3090. The average training time for a single epoch is 40 seconds on SMM4H and 90 seconds on CADEC for the base models. The training time increases slightly for the architectures using the additional LSTM layer and doubles for the architectures using the CRF layer.


\section{Results and Discussion}
\label{sec:eval}

First, we start by analyzing the performance of the base Transformer-based architectures without additional processing modules.
We discuss the results of all the models on the two datasets, taking into account the \shapvals to discover patterns in the performances of the models.
Secondly, we discuss the effects of using the additional CRF and LSTM modules, and how they have different effects on the two datasets.

\subsection{Base Models Performance}


\begin{figure}[!htbp]
\centering
\includegraphics[width=.9\linewidth]{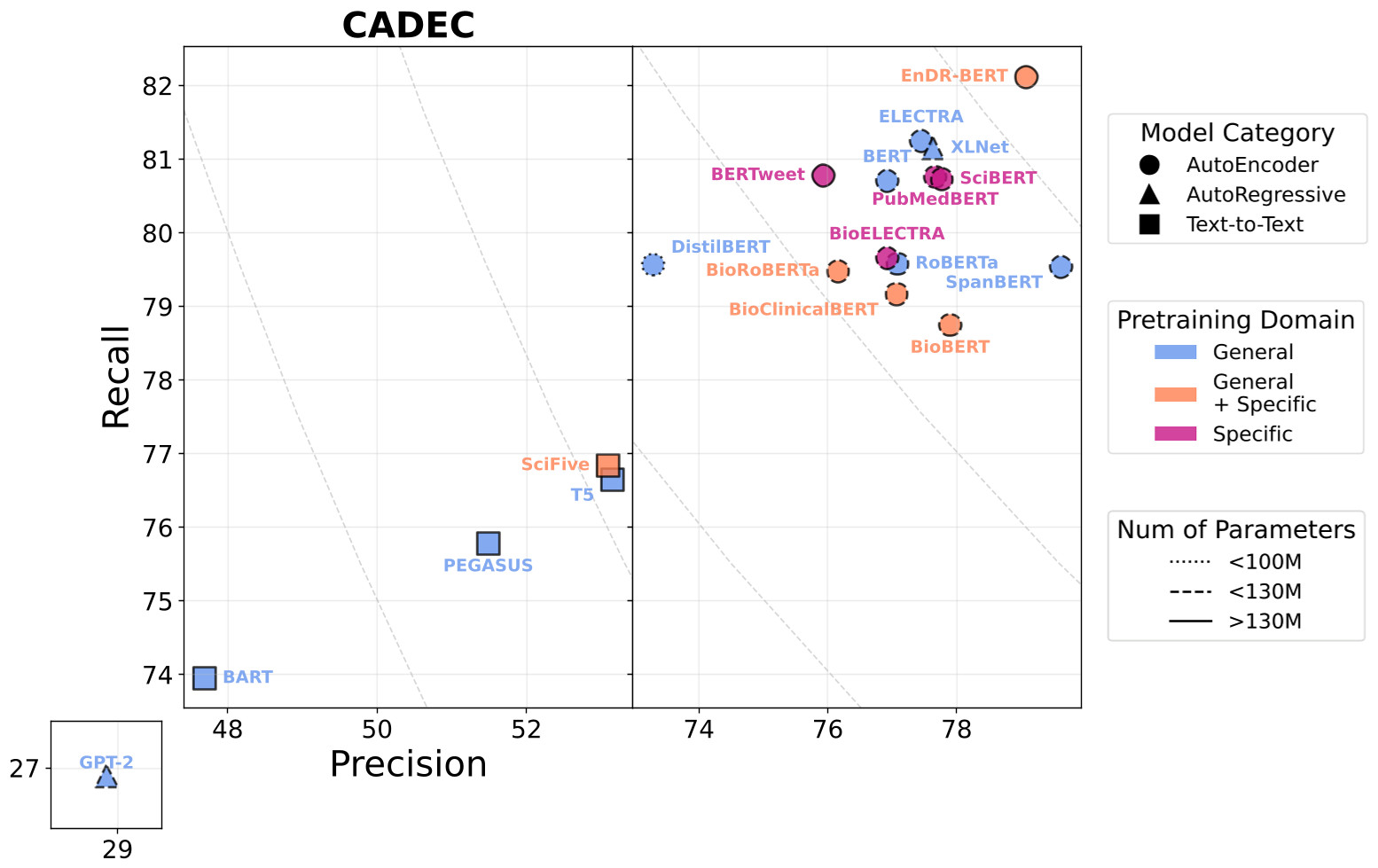}
\caption{Precision and Recall of all the base models (with no additional modules) on the CADEC dataset.}
\label{fig:precision_recall_cadec}
\end{figure}

\begin{figure}[!htbp]
\centering
\includegraphics[width=.9\linewidth]{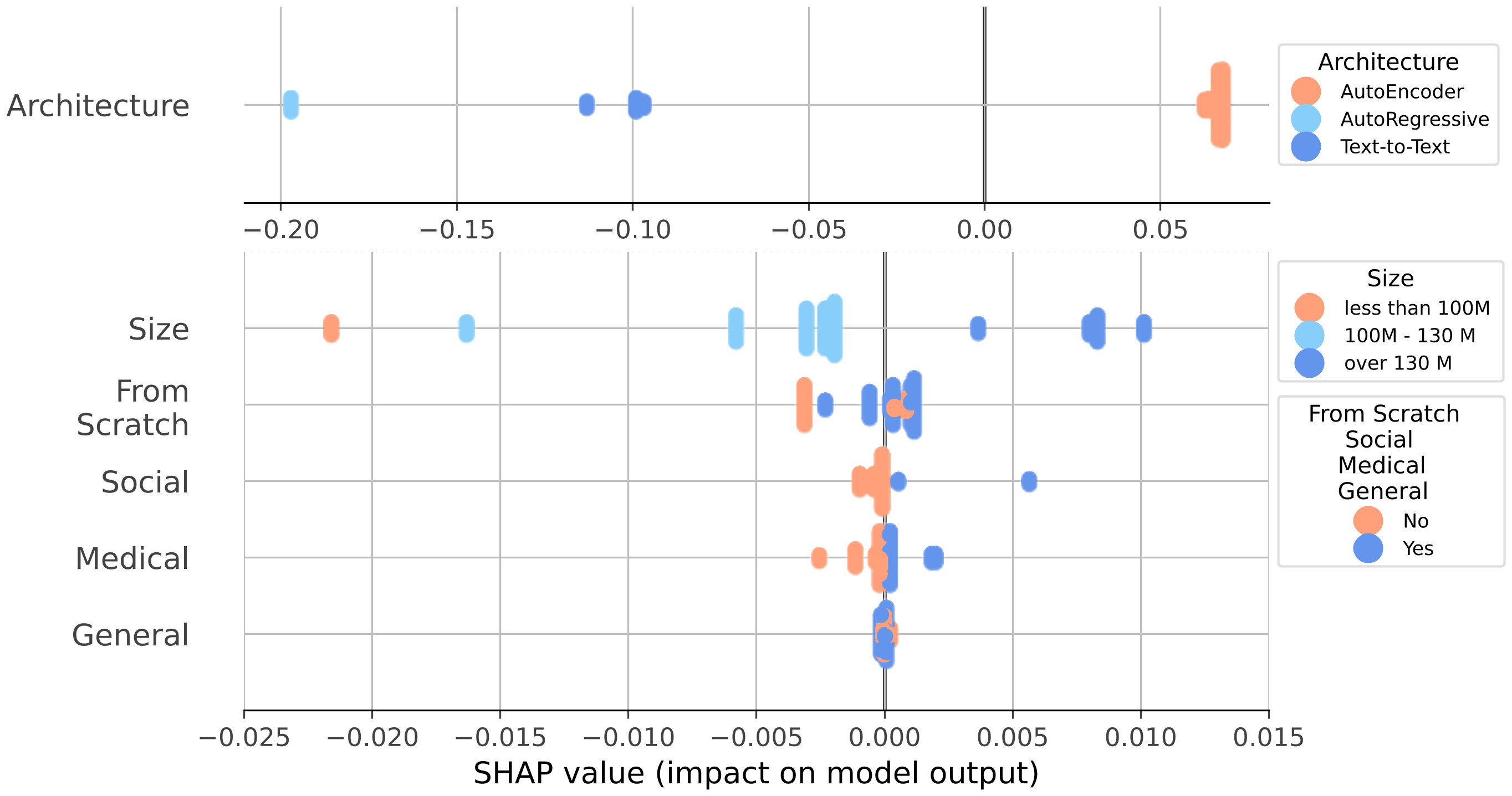}
\caption{\shapvals calculated on the CADEC dataset.}
\label{fig:shap_cadec}
\end{figure}

We start by analyzing the performance of all the base models (without additional LSTM/CRF modules) on the CADEC dataset. We report the Precision and Recall of the models in Figure \ref{fig:precision_recall_cadec}, while Figure \ref{fig:shap_cadec} contains the results of the feature importance analysis performed with the \shapvals.

In Figure \ref{fig:precision_recall_cadec}, different shapes represent different models categories: \autoencodershape for \autoencoders, \autoregressiveshape for \autoregressive models, and \texttotextshape for \texttotext models. The colors show the domain of the training data of the model: general in \generaldomaincolor \generaldomain, specialized (medical or social) in \indomaincolor \indomain, mixed general and specialized in \mixeddomaincolor \mixeddomain.
The linestyle of the markers shows the size of the model: dotted if the model has less than 100M parameters, dashed if the model has between 100M and 130M parameters, and solid if it is larger than 130M parameters.
The dashed gray lines on the plot are iso-F1 curves, showing points of equal F1-score.

As regards Figure \ref{fig:shap_cadec}, each row represents one of the features used by the Random Forest to predict the F1 score of the models.
Each point in a row is a sample, and its color represents the input value of its feature. For example, considering the feature \textit{Architecture}, coral points are \autoencoders, light-blue points are \autoregressive models, and blue points are \texttotext models.
The x-coordinate represents SHAP values, which are positive if the feature contributes to a higher F1 score, and negative if it decreases it.
The features are arranged in order of importance, from top to bottom, based on the magnitude of the SHAP values (i.e., their impact on the F1 score). 

\medskip

Looking at Figure \ref{fig:precision_recall_cadec}, we can clearly distinguish three clusters of models: the \autoencoders \autoencodershape in the top right (together with \xlnet), which reach the highest performance; the \texttotext models \texttotextshape, which have a considerably lower Precision; \gpttwo (one of the \autoregressive models \autoregressiveshape), which has the worst performance overall and is clearly separated from the other Transformer variants.
This is confirmed by the \shapvals (Figure \ref{fig:shap_cadec}), which show that \textit{Architecture} is the most impactful feature, and its three values (coral, light-blue and blue) have different impacts (negative or positive) on the expected F1 score.
All the models based on text generation (except \xlnet) have a low Recall (lower than 77\%), and even lower Precision (lower than 53\%), which clearly separates them from the \autoencoders. The low Recall is probably caused by the high number of ADEs that need to be generated for the CADEC dataset, as the \texttotext models seem to struggle to generate long sequences of ADEs.

If we focus on the cluster of \autoencoders, we can see that the best-performing model overall is \xyendr, which is also the largest \autoencoder model (solid outline). Conversely, the worst model of the cluster is \distilbert, which is the smallest model (dotted outline). Smaller models generally lead to a lower F1 score, which is also attested by the \shapvals(\textit{Size} is the second most impactful feature on the performance of the models).

The third most impactful feature on Figure \ref{fig:shap_cadec} is \textit{From scratch}: models which are not pre-trained from scratch have generally a lower performance compared with the ones pre-trained from scratch. These models correspond to the five mixed-domain ones \mixeddomain and \distilbert. We can see that this is mostly true for the three \autoencoders \bioroberta, \bioclinicalbert, and \biobert. \xyendr and \scifive counter the relative drop in performance thanks to their large size.

Another interesting observation from the \shapvals, is that pre-training on \textit{Social} or \textit{Medical} data has a positive impact on the model's performance (blue points have positive SHAP values). These models correspond to the in-domain models \indomain in Figure \ref{fig:precision_recall_cadec}, where they are shown to achieve the same performance as models trained on general-domain data \generaldomain. The positive contribution seems to be too small to have an effect on this plot, where it is overshadowed by the effects of the other model characteristics.

Overall, the model that achieves the highest Precision on CADEC is \spanbert (general-domain \generaldomain), while the one with the highest Recall and F1 score is \xyendr. Since the texts and ADE mentions present in CADEC are particularly long (see Character Count in Table \ref{tab:textstats}), \spanbert probably has an advantage over other models thanks to its span-based pre-training.

\begin{figure}[!htbp]
\centering
\includegraphics[width=.9\linewidth]{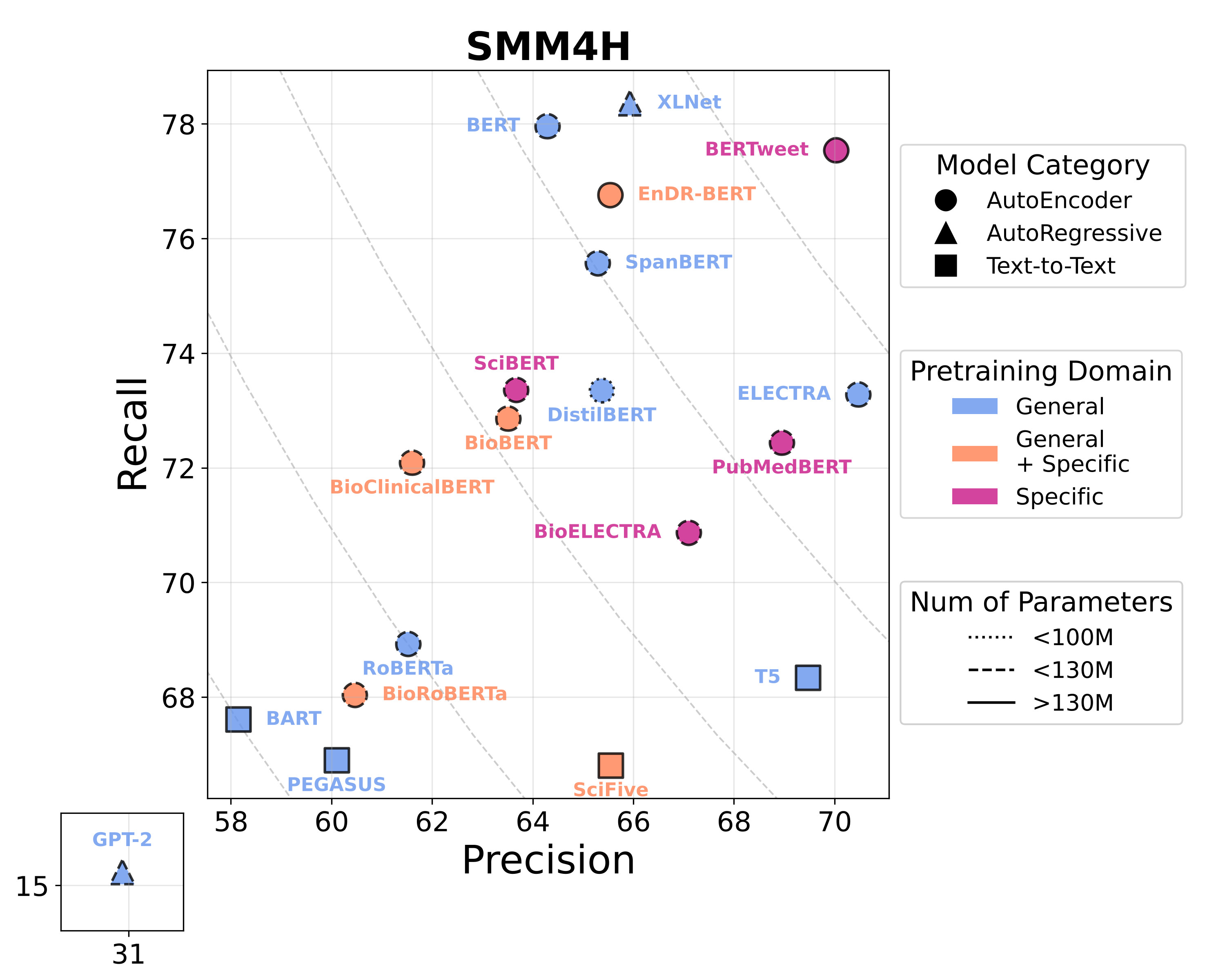}
\caption{Precision and Recall of all the base models (with no additional modules) on the SMM4H dataset.}
\label{fig:precision_recall_smm}
\end{figure}

\begin{figure}[!htbp]
\centering
\includegraphics[width=.9\linewidth]{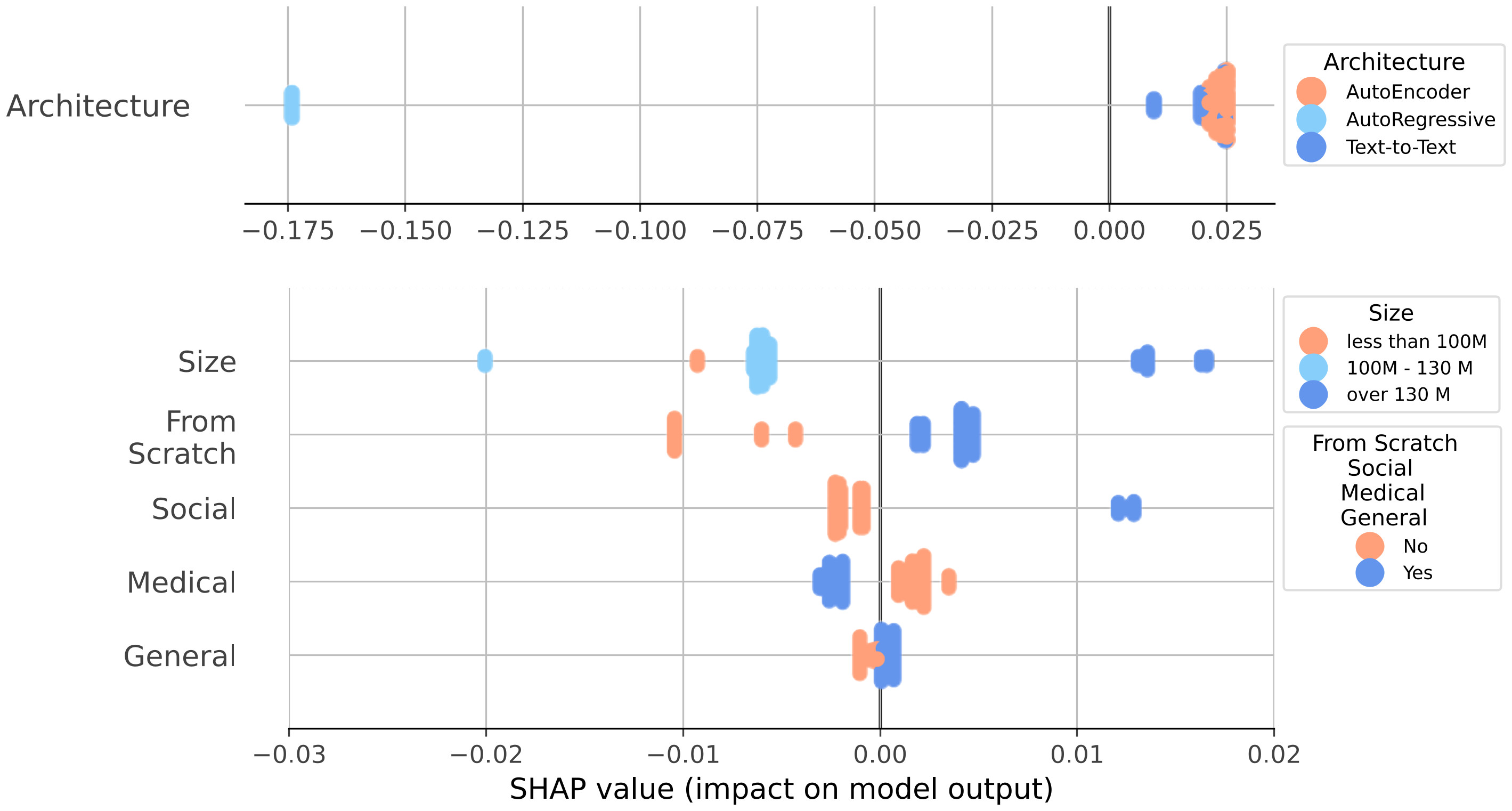}
\caption{\shapvals calculated on the SMM4H dataset.}
\label{fig:shap_smm}
\end{figure}

\medskip

Figure \ref{fig:precision_recall_smm} and \ref{fig:shap_smm} report the same information as the previous ones, but for the SMM4H dataset. The order of the most impactful features according to the \shapvals is the same for the two datasets.

Differently from CADEC, there are two noticeable clusters in Figure \ref{fig:precision_recall_smm}: \gpttwo and all the other Transformer-based variants. Similarly to CADEC, the \texttotext models \texttotextshape have a lower Recall than most of the \autoencoders \autoencodershape. However, on the SMM4H dataset their Recall is closer to the other \autoencoders (\roberta and \bioroberta), and their Precision is also on-par with most of the other ones. For these reasons, they do not create a separate performance cluster as happened in CADEC. 
This is further confirmed by the \shapvals in Figure \ref{fig:shap_smm}: the feature \textit{Architecture} presents only two clusters (\autoregressive vs others), and both \autoencoder (coral) and \texttotext (blue) samples contribute to an increase in F1 score.

The model \textit{Size} is still the second most impactful SHAP feature, but its effect on the Precision-Recall plot is more difficult to observe.

In sharp contrast with CADEC, the use of \textit{Medical} domain pre-training leads to a lower performance, while \textit{General} domain data slightly increases it. \textit{Social} data pre-training also has a sharp positive impact. This indicates that medical in-domain knowledge leads to no advantage when dealing with highly informal texts, such as tweets. 
Indeed, most of the models that reach the best performance in terms of Precision, Recall, or F1-score are trained on general-domain data only (e.g., \xlnet, \bert, and \electra). The only cases in which in-domain knowledge brings an advantage are \bertweet and \xyendr, which are trained on social media texts (tweets and forum posts), highlighting that, in this case, social media pre-training is more valuable than medical knowledge.

Finally, the effect of training a model \textit{From Scratch} is more noticeable on SMM4H, where it leads to a small increase in performance. Models that are not trained from scratch correspond to the mixed-domain models \mixeddomain in Figure \ref{fig:precision_recall_smm}, and this decreases their F1 score according to the \shapvals. This loss in performance is probably connected to the fact that most mixed-domain models include medical knowledge, which is not beneficial on the SMM4H dataset.

\medskip

To summarize, for both datasets: text-generation models (\autoregressive and \texttotext) lead to the lowest performance; larger models tend to have higher performance; using models trained from scratch (regardless of their domain) is beneficial; knowledge of social media language is highly beneficial.

The main difference between the two datasets is that models pre-trained on medical data have lower performance on SMM4H, due to the large gap in textual style.

\subsection{Error Analysis}

Given the large number of analyzed models, it is challenging to perform an in-depth error analysis and compare the kind of errors produced by the various base models. However, we performed a qualitative analysis of the output of the models as follows. We fixed one of the thirty random seeds and gathered all the predictions of the 19 base models. We divided the predictions into the following sets: Correct, Partial, Missing and Spurious (see Section \ref{sec:metrics} for the definitions). We then compared these sets of predictions for all the models, creating rankings (ordering the predictions according to how many models classified it as Correct/Missing/Spurious) and grouping the predictions by topic (e.g., sleep disorders or weight change). The following trends emerged:

\begin{itemize}
\item \textit{Spurious predictions on CADEC and SMM4H}. For both of the datasets, 80\% of the Spurious predictions are unique (predicted only once and by less than three models out of 19). The spurious entities which are wrongly extracted/generated by all the models are short one-word entities, which are diseases (or symptoms of a disease, such as \quot{headaches}), but denote real ADE mentions in other samples.
\item \textit{\autoregressive and \texttotext models on CADEC}. A large amount of the gold entities belongs to the Missing set for all the models and are never predicted (neither as Correct nor as Partial). The entities which are consistently Missing for all the text-generation models are composed of multiple words (e.g., \quot{affected my balance}, \quot{blood pressure elevated}, \quot{altered my heart function}) and they are often very technical (e.g., \quot{peripheral neuropathy}, \quot{gastrointestinal cramping}, \quot{rheumatoid arthrtitis}).\\ On the other hand, the entities which are predicted correctly by all the text-generation models (Correct and Partial) are short, one-word entities which are present in multiple samples (e.g., \quot{constipation}, \quot{diarrhea}, \quot{fatigue}, \quot{insomnia}).
\item \textit{\autoencoder models on CADEC}. The proportion of Missing entities for the \autoencoder models is significantly smaller, as confirmed by their higher Recall. The entities which are missed by all the models are extremely short ones (e.g., \quot{sick}, \quot{pain}, \quot{gas}), which are difficult to contextualize, and extremely long ones (e.g., \quot{will never get back the full use of my arms or legs}). In general, all models struggle to identify ADE with long character counts. 
\item \textit{All models on SMM4H}. The overall number of Missing entities is low. The ones which are shared among all the models are extremely short, and some of them are hashtags (e.g., \quot{\#nosleepp}, \quot{\#wideawake}). The Missing entities which are common for all the models trained on medical domain only are short, colloquial terms such as \quot{puking}, \quot{out of it} and \quot{passing out}.
\end{itemize}

\begin{figure}[!htbp]
\centering
\includegraphics[width=.95\linewidth]{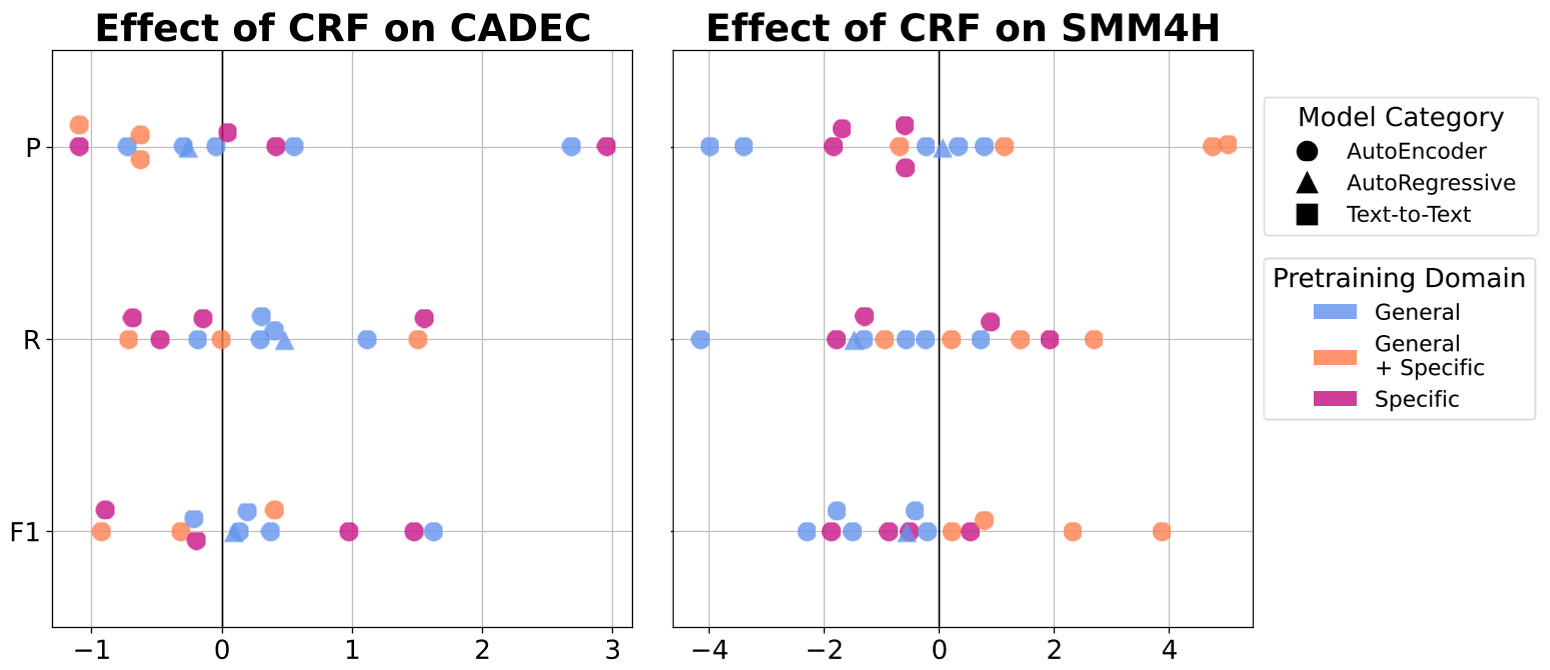}
\caption{Effect of the CRF module on Precision and Recall.}
\label{fig:precision_recall_crf}
\end{figure}

\subsection{Effect of the CRF}

Figure \ref{fig:precision_recall_crf} shows the effect that using the additional CRF module has on the \autoencoder model and \xlnet. The plots report the difference between the metric achieved by the model with the CRF module and the one without. Positive values indicate an advantage in using the additional module, while negative values mean it decreases the base performance of the model.

Looking at the results on the CADEC dataset, we observe that the CRF module generally has a positive impact on the Recall of the models and a mixed impact on the Precision.
It leads to a gain of up to 3 points in Precision and up to 1.5 points in Recall, leading to an overall increase in F1-score too.
There seems to be no pattern that relates the pre-training domain with the effect of the CRF module.

On the SMM4H dataset, the CRF module seems to have different effects based on the pre-training domain of the models.
It leads to a decrease in Precision for models with specific or general-domain knowledge (\indomain \generaldomain), with a subsequent loss in F1-score.
On the contrary, mixed-domain models (\mixeddomain) experience a gain in Precision (up to 4 points) and in Recall, with an overall increase in F1-score.

\begin{figure}[!htbp]
\centering
\includegraphics[width=.95\linewidth]{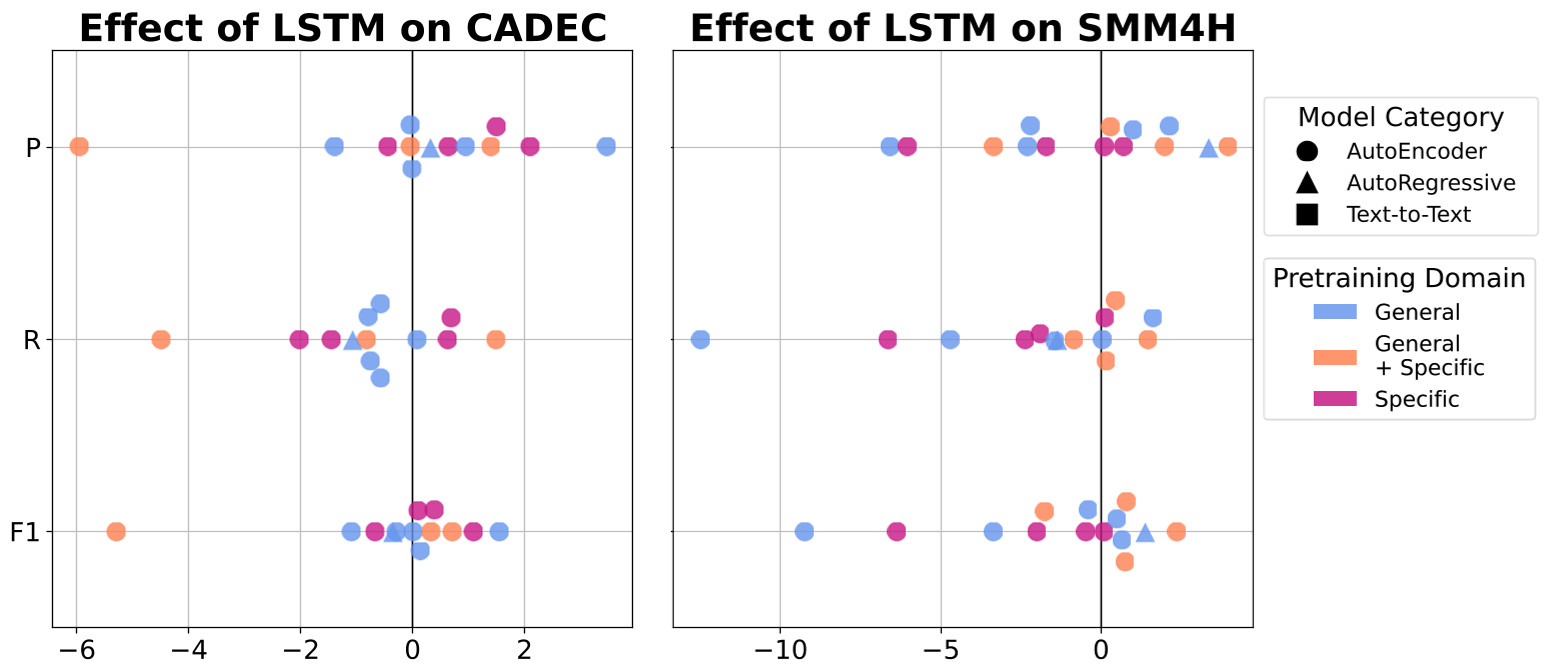}
\caption{Effect of the LSTM module on Precision and Recall.}
\label{fig:precision_recall_lstm}
\end{figure}

\subsection{Effect of the LSTM}

In Figure \ref{fig:precision_recall_lstm} we report the results for the LSTM module using the same format of Figure \ref{fig:precision_recall_crf}.

On the CADEC dataset, the LSTM generally increases the Precision of the models (up to 2.5 points) and has a small (mostly negative) impact on the Recall, which is more frequent for general-domain (\generaldomain) models. The overall effect is generally an increase in F1-score.

The effect of the LSTM on the SMM4H dataset seems to show no regularities: the Precision increases or decreases with no definite pattern, almost all the models experience a drop in Recall (up to 12.5 points). The overall effect on the F1-score is negative.

The LSTM seems to have a similar effect on both datasets, therefore it could be reasonable to use it in cases where we are interested in increasing the Precision of the base model.

\subsection{Take-home Messages}

To summarize the results of all the previous experiments, we observed that:

\begin{itemize}

\item \autoencoder models are the best choice of model to deal with this task, while models based on text generation (\autoregressive and \texttotext) do not have good performances on long texts or texts that contain a high number of ADEs;

\item when all other features are the same, bigger models have the highest performance on both formal and informal texts;

\item pre-training on social media texts leads to a consistent increase in performance on both datasets, while medical pre-training is only effective when working with social media texts that have a more formal language (in this case the forum posts from CADEC);

\item the use of additional modules needs to be evaluated on a case-by-case basis. On the whole, the CRF module has a positive impact on the Recall of the models when used in formal texts, and positive effects on Precision for mixed-domain models in informal texts. On the other hand, the LSTM tends to increase the Precision of the models in formal texts but has negative effects on most of the metrics in informal texts.

\end{itemize}

\section{Conclusions}

In this paper, we performed a systematic analysis of 19 transformer-based models for ADE extraction on informal texts. We compared their performance on two datasets with different textual styles, and correlated it with the following model features: category (\autoencoder, \autoregressive, \texttotext), pre-training domain, training from scratch, and model size in number of parameters.
We used feature importance techniques to correlate each of these characteristics to the performance of the models.
Furthermore, we analyzed the impact of commonly-used additional processing layers (CRF and LSTM) on the performance of the models.
To conclude our analyses, we presented a list of take-home messages that can be derived from the experimental data.

Since the code we used for these analyses is publicly available, it will be possible to adapt it and use it for other tasks. In particular, future researchers will be able to use it to test different kinds of models comparing their features and performances on other tasks and domains.

In the future, we plan to expand our analyses to other tasks in the medical domain.
This will help building a more solid understanding of which model characteristics are more effective for each task, especially in the new field of digital pharmacovigilance on social media texts. 



\appendix

\section{Further details on the models}
\label{app:hyperparam}

Table \ref{tab:model_urls} reports the unique identifiers of all the models in the HuggingFace library for reproducibility.

\begin{table*}[!hbtp]
\centering
\resizebox{\linewidth}{!}{%
\begin{tabular}{rl}
\hline
\textbf{Name} & \textbf{Model name in the Transformers library} \\ \hline
\bert            &  \texttt{bert-base-uncased}\\
\distilbert      &  \texttt{distilbert-base-uncased}\\
\spanbert        &  \texttt{SpanBERT/spanbert-base-cased} \\
\roberta         &  \texttt{roberta-base}\\
\electra         &  \texttt{google/electra-base-discriminator}\\
\xlnet           &  \texttt{xlnet-base-cased}\\
\gpttwo          &  \texttt{gpt2}\\
\tfive           &  \texttt{t5-base}\\
\pegasus         &  \texttt{google/pegasus-xsum}\\
\bart            &  \texttt{facebook/bart-base}\\ \hline
\bertweet        &  \texttt{vinai/bertweet-large} \\
\biobert         &  \texttt{monologg/biobert\_v1.1\_pubmed} \\
\bioclinicalbert &  \texttt{emilyalsentzer/Bio\_ClinicalBERT}\\
\scibert         &  \texttt{allenai/scibert\_scivocab\_cased} \\
\pubmedbert      &  \texttt{microsoft/BiomedNLP-PubMedBERT-base-uncased-abstract-fulltext}\\
\xyendr          &  \texttt{cimm-kzn/endr-bert}\\
\bioelectra      &  \texttt{kamalkraj/bioelectra-base-discriminator-pubmed}\\
\bioroberta      &  \texttt{allenai/biomed\_roberta\_base}\\
\scifive         &  \texttt{razent/SciFive-base-Pubmed}\\ \hline
\end{tabular}
}
\caption{Unique identifiers of the models in the Huggingface Transformer library.}
\label{tab:model_urls}
\end{table*}

Table \ref{tab:best_params_bert_wrapper} contains
the best hyperparameters used for all architectures on SMM4H and CADEC, respectively.

\begin{table}[!hbtp]
\centering
\resizebox{0.67\linewidth}{!}{%

\begin{tabular}
{r c@{$\quad$}l@{$\quad$}r@{$\quad$}c@{$\quad$} c@{$\quad$}l@{$\quad$}r@{$\quad$}c@{$\quad$}}

\hline

& \multicolumn{4}{c}{\textbf{CADEC}} & \multicolumn{4}{c}{\textbf{SMM4H20}}\\
\textbf{Model Name} & \rotatebox{90}{\textbf{lr}} & \rotatebox{90}{\textbf{dropout}} & \rotatebox{90}{\textbf{epoch}} & \rotatebox{90}{\textbf{batch\_size}} & \rotatebox{90}{\textbf{lr}} & \rotatebox{90}{\textbf{dropout}} & \rotatebox{90}{\textbf{epoch}} & \rotatebox{90}{\textbf{batch\_size}}\\

\hline

                    \bert   &     1e-4 &    0.25 &      6 &   4 &   5e-5 &    0.25 &     10 &   16\\
                \bertweet   &     5e-5 &     0.3 &      8 &   4 &   5e-5 &    0.15 &      7 &   16\\
                 \biobert   &     1e-4 &     0.2 &      7 &   8 &   5e-5 &     0.2 &      5 &   16\\
         \bioclinicalbert   &     5e-5 &     0.2 &      8 &   4 &   5e-5 &    0.15 &      3 &    8\\
              \bioelectra   &     5e-5 &     0.2 &      9 &   8 &   1e-4 &     0.2 &      4 &    8\\
              \bioroberta   &     1e-4 &    0.25 &     15 &   8 &   5e-5 &    0.15 &      7 &   16\\
              \distilbert   &     5e-5 &     0.3 &      7 &   8 &   1e-4 &     0.3 &      4 &    8\\
                 \electra   &     5e-5 &    0.15 &      8 &   8 &   5e-5 &    0.15 &      7 &   32\\
                \xyendr     &     5e-5 &     0.3 &     14 &   4 &   5e-5 &    0.15 &      5 &   32\\
              \pubmedbert   &     5e-5 &     0.3 &     10 &   4 &   5e-5 &     0.2 &     10 &   16\\
                 \roberta   &     5e-5 &    0.15 &     10 &   8 &   5e-5 &    0.15 &      7 &    8\\
                 \scibert   &     5e-5 &     0.3 &     13 &   4 &   5e-5 &     0.3 &     13 &    8\\
                \spanbert   &     5e-5 &    0.15 &     10 &   4 &   5e-5 &    0.15 &      8 &    8\\
                   \xlnet   &     5e-5 &    0.15 &      7 &   4 &   5e-5 &     0.2 &     15 &   32\\
                \tfive      &     2-e4 &    0.15 &     9 &    4 &   5e-5 &    0.15 &     10 &    8 \\
                \gpttwo     &     1-e3 &    0.15 &     6 &    8 &   5e-5 &    0.15 &      4 &   32 \\
                     \bart  &     5-e5 &    0.15 &    10 &   32 &   6e-5 &    0.15 &     10 &   16 \\
                  \pegasus  &     2-e4 &    0.15 &     5 &    4 &   5e-5 &    0.15 &      8 &    8 \\
                  \scifive  &     6-e5 &    0.15 &    12 &    4 &   1e-4 &    0.15 &     11 &    8 \\
\hline
                \bert+CRF   &     1e-4 &     0.3 &      9 &   4 &   1e-4 &     0.3 &      7 &   16\\
            \bertweet+CRF   &     5e-5 &    0.25 &     11 &   4 &   5e-5 &    0.15 &      6 &   32\\
             \biobert+CRF   &     5e-5 &    0.25 &      7 &   8 &   1e-4 &     0.2 &      6 &    8\\
     \bioclinicalbert+CRF   &     5e-5 &    0.25 &      7 &   4 &   1e-4 &     0.2 &      5 &   16\\
          \bioelectra+CRF   &     5e-5 &     0.3 &     15 &   4 &   1e-4 &    0.15 &      5 &   16\\
          \bioroberta+CRF   &     5e-5 &    0.25 &      9 &   4 &   5e-5 &    0.15 &      8 &   16\\
          \distilbert+CRF   &     1e-4 &     0.3 &      8 &   8 &   1e-4 &    0.25 &      5 &    8\\
             \electra+CRF   &     5e-5 &    0.25 &     10 &   4 &   1e-4 &    0.25 &      7 &   16\\
            \xyendr+CRF     &     1e-4 &    0.25 &      8 &   8 &   5e-5 &    0.15 &      4 &   16\\
          \pubmedbert+CRF   &     1e-4 &     0.3 &     14 &   8 &   5e-5 &     0.2 &      8 &    8\\
             \roberta+CRF   &     5e-5 &    0.15 &      8 &   4 &   5e-5 &     0.2 &     10 &    8\\
             \scibert+CRF   &     5e-5 &     0.2 &      6 &   4 &   1e-4 &     0.2 &      6 &   32\\
            \spanbert+CRF   &     5e-5 &    0.15 &      9 &   8 &   5e-5 &    0.15 &      7 &    8\\
               \xlnet+CRF   &     5e-5 &    0.25 &     12 &   4 &   5e-5 &    0.15 &      7 &   16\\
\hline
               \bert+LSTM   &     5e-5 &     0.2 &      7 &   4 &   5e-5 &    0.25 &      7 &    8\\
           \bertweet+LSTM   &     5e-5 &    0.15 &      8 &   4 &   5e-5 &    0.15 &     15 &   32\\
            \biobert+LSTM   &     5e-5 &     0.2 &      8 &   4 &   5e-5 &    0.25 &      7 &    8\\
    \bioclinicalbert+LSTM   &     1e-4 &     0.3 &      9 &   4 &   5e-5 &    0.25 &      9 &   16\\
         \bioelectra+LSTM   &     5e-5 &    0.25 &     12 &   4 &   5e-5 &     0.2 &      8 &    8\\
         \bioroberta+LSTM   &     5e-5 &    0.15 &      9 &   4 &   5e-5 &    0.25 &     14 &    8\\
         \distilbert+LSTM   &     1e-4 &     0.2 &      6 &   4 &   1e-4 &     0.2 &      6 &   16\\
            \electra+LSTM   &     5e-5 &     0.2 &      6 &   4 &   5e-5 &     0.2 &     11 &   32\\
           \xyendr+LSTM     &     1e-4 &     0.3 &      9 &   8 &   5e-5 &    0.15 &      5 &    8\\
         \pubmedbert+LSTM   &     1e-4 &    0.15 &      8 &   8 &   5e-5 &    0.15 &      8 &   16\\
            \roberta+LSTM   &     5e-5 &    0.15 &     10 &   4 &   5e-5 &     0.2 &     14 &   16\\
            \scibert+LSTM   &     5e-5 &     0.3 &     12 &   4 &   5e-5 &    0.25 &     14 &   32\\
           \spanbert+LSTM   &     5e-5 &    0.15 &      8 &   8 &   5e-5 &     0.2 &     10 &    8\\
              \xlnet+LSTM   &     5e-5 &    0.15 &      9 &   4 &   5e-5 &    0.15 &     12 &   32\\
\hline
\end{tabular}
}
\caption{Best hyperparameters for all models.}
\label{tab:best_params_bert_wrapper}
\end{table}

\section{Detailed metrics of all the models}
\label{app:full_metrics}

The following tables report the Strict and Relaxed evaluation metrics for all the models used in the paper. Tables \ref{tab:bert_wrapper_smm4h20}--\ref{tab:bert_lstm_smm4h20} report the results on SMM4H. Tables \ref{tab:bert_wrapper_cadec}--\ref{tab:bert_lstm_cadec} report the results on CADEC.

\begin{table*}[!hbtp]
\centering
\resizebox{\linewidth}{!}{%
\begin{tabular}{r ccc ccc}
\hline
& \multicolumn{3}{c}{\textbf{Relaxed}} & \multicolumn{3}{c}{\textbf{Strict}}\\
& \textbf{F1} & \textbf{P} & \textbf{R} & \textbf{F1} & \textbf{P} & \textbf{R}\\
\hline
\bert & \asd{70.43}{0.22} & \asd{64.29}{1.53} & \asd{77.96}{1.79} & \asd{61.99}{0.90} & \asd{56.65}{1.44} & \asd{68.52}{2.11}\\
\distilbert & \asd{69.09}{1.36} & \asd{65.37}{2.62} & \asd{73.35}{1.27} & \asd{60.70}{1.41} & \asd{57.43}{2.39} & \asd{64.44}{1.48}\\
\spanbert & \asd{70.04}{1.01} & \asd{65.29}{1.76} & \asd{75.57}{0.10} & \asd{62.59}{1.39} & \asd{57.84}{2.83} & \asd{68.33}{0.85}\\
\roberta & \asd{64.97}{1.08} & \asd{61.52}{1.95} & \asd{68.93}{2.14} & \asd{56.12}{1.23} & \asd{53.25}{1.99} & \asd{59.41}{1.86}\\
\electra & \asd{71.81}{1.51} & \asd{70.47}{1.89} & \asd{73.28}{2.60} & \asd{63.46}{1.91} & \asd{62.45}{1.73} & \asd{64.58}{3.09}\\
\xlnet & \asd{71.55}{0.52} & \asd{65.93}{1.79} & \asd{78.36}{2.37} & \asd{62.89}{0.88} & \asd{57.80}{1.23} & \asd{69.07}{2.71}\\
\gpttwo & \asd{20.15}{2.74} & \asd{30.83}{2.58} & \asd{15.19}{2.85} & \asd{11.73}{4.74} & \asd{17.16}{5.87} & \asd{08.97}{3.82}\\
\tfive & \asd{68.90}{1.08} & \asd{69.47}{1.60} & \asd{68.34}{0.77} & \asd{61.90}{1.08} & \asd{62.42}{1.58} & \asd{61.40}{0.72}\\
\pegasus & \asd{63.31}{0.84} & \asd{60.10}{1.38} & \asd{66.90}{0.76} & \asd{55.90}{1.05} & \asd{53.07}{1.47} & \asd{59.07}{0.89}\\
\bart & \asd{62.44}{1.81} & \asd{58.15}{3.87} & \asd{67.61}{1.42} & \asd{54.35}{1.85} & \asd{50.62}{3.61} & \asd{58.85}{1.11}\\
\hline
\bertweet & \asd{73.57}{0.72} & \asd{70.03}{1.07} & \asd{77.54}{2.17} & \asd{64.44}{1.56} & \asd{61.34}{1.38} & \asd{67.93}{2.69}\\
\biobert & \asd{67.83}{0.72} & \asd{63.51}{1.56} & \asd{72.86}{2.03} & \asd{59.62}{1.56} & \asd{55.81}{1.47} & \asd{64.06}{2.81}\\
\bioclinicalbert & \asd{66.42}{1.19} & \asd{61.60}{1.73} & \asd{72.09}{1.16} & \asd{57.52}{1.20} & \asd{53.40}{1.62} & \asd{62.36}{1.26}\\
\scibert & \asd{68.14}{0.72} & \asd{63.67}{1.96} & \asd{73.36}{1.40} & \asd{59.77}{0.94} & \asd{55.91}{1.87} & \asd{64.28}{1.34}\\
\pubmedbert & \asd{70.63}{0.91} & \asd{68.95}{1.13} & \asd{72.44}{1.82} & \asd{63.00}{1.18} & \asd{61.90}{2.00} & \asd{64.23}{2.15}\\
\xyendr & \asd{70.64}{1.21} & \asd{65.54}{2.88} & \asd{76.76}{1.38} & \asd{62.36}{1.33} & \asd{57.39}{2.26} & \asd{68.37}{1.32}\\
\bioroberta & \asd{64.01}{0.83} & \asd{60.46}{0.71} & \asd{68.04}{1.90} & \asd{54.68}{1.20} & \asd{50.85}{1.55} & \asd{59.24}{2.60}\\
\bioelectra & \asd{68.93}{1.40} & \asd{67.10}{1.54} & \asd{70.87}{1.75} & \asd{61.62}{1.78} & \asd{59.99}{1.84} & \asd{63.36}{2.04}\\
\scifive & \asd{66.16}{1.09} & \asd{65.55}{2.01} & \asd{66.81}{1.02} & \asd{59.75}{1.14} & \asd{59.20}{2.05} & \asd{60.34}{0.74}\\
\hline
\end{tabular}
}
\caption{Metrics of all the base models on SMM4H.}
\label{tab:bert_wrapper_smm4h20}
\end{table*}

\begin{table*}[!hbtp]
\centering
\resizebox{\linewidth}{!}{%
\begin{tabular}{r ccc ccc}
\hline
& \multicolumn{3}{c}{\textbf{Relaxed}} & \multicolumn{3}{c}{\textbf{Strict}}\\
& \textbf{F1} & \textbf{P} & \textbf{R} & \textbf{F1} & \textbf{P} & \textbf{R}\\
\hline
\bert & \asd{70.19}{0.42} & \asd{64.03}{1.07} & \asd{77.69}{1.10} & \asd{62.14}{0.62} & \asd{56.69}{0.90} & \asd{68.79}{1.34}\\
\distilbert & \asd{68.64}{0.52} & \asd{65.67}{1.82} & \asd{72.01}{1.99} & \asd{59.19}{1.09} & \asd{57.14}{1.41} & \asd{61.42}{1.29}\\
\spanbert & \asd{68.50}{2.99} & \asd{66.04}{3.68} & \asd{71.39}{4.50} & \asd{60.25}{3.71} & \asd{58.10}{4.21} & \asd{62.79}{4.73}\\
\roberta & \asd{62.64}{3.46} & \asd{58.09}{5.54} & \asd{68.32}{0.99} & \asd{53.38}{4.78} & \asd{49.46}{6.12} & \asd{58.21}{2.76}\\
\electra & \asd{70.00}{1.61} & \asd{66.45}{1.43} & \asd{73.97}{1.82} & \asd{61.32}{2.68} & \asd{56.25}{3.82} & \asd{67.49}{0.97}\\
\xlnet & \asd{70.97}{1.16} & \asd{65.97}{1.85} & \asd{76.86}{1.60} & \asd{61.17}{1.55} & \asd{56.81}{2.03} & \asd{66.31}{1.80}\\
\hline
\bertweet & \asd{74.08}{0.96} & \asd{69.41}{1.45} & \asd{79.43}{0.64} & \asd{65.24}{1.19} & \asd{61.13}{1.31} & \asd{69.96}{1.40}\\
\biobert & \asd{70.12}{1.81} & \asd{68.50}{1.29} & \asd{71.88}{3.10} & \asd{63.25}{2.94} & \asd{61.77}{2.13} & \asd{64.86}{4.10}\\
\bioclinicalbert & \asd{70.26}{1.24} & \asd{66.32}{1.67} & \asd{74.75}{2.02} & \asd{62.25}{2.33} & \asd{58.35}{2.70} & \asd{66.78}{2.77}\\
\scibert & \asd{67.23}{0.92} & \asd{63.04}{1.08} & \asd{72.03}{1.02} & \asd{58.62}{1.04} & \asd{54.96}{1.16} & \asd{62.81}{1.09}\\
\pubmedbert & \asd{70.08}{1.14} & \asd{67.23}{2.14} & \asd{73.30}{2.51} & \asd{62.65}{1.36} & \asd{60.20}{2.19} & \asd{65.42}{2.38}\\
\xyendr & \asd{71.39}{1.09} & \asd{66.64}{1.93} & \asd{76.94}{1.67} & \asd{63.44}{1.18} & \asd{58.88}{2.35} & \asd{68.88}{1.62}\\
\bioroberta & \asd{64.20}{1.09} & \asd{59.74}{0.64} & \asd{69.42}{2.32} & \asd{54.94}{0.86} & \asd{50.90}{0.58} & \asd{59.73}{2.16}\\
\bioelectra & \asd{67.02}{1.91} & \asd{65.23}{2.13} & \asd{69.05}{3.49} & \asd{59.02}{1.90} & \asd{58.20}{2.76} & \asd{59.93}{1.96}\\
\hline
\end{tabular}
}
\caption{Metrics of the \autoencoder models with CRF module on SMM4H.}
\label{tab:bert_crf_smm4h20}
\end{table*}

\begin{table*}[!hbtp]
\centering
\resizebox{\linewidth}{!}{%
\begin{tabular}{r ccc ccc}
\hline
& \multicolumn{3}{c}{\textbf{Relaxed}} & \multicolumn{3}{c}{\textbf{Strict}}\\
& \textbf{F1} & \textbf{P} & \textbf{R} & \textbf{F1} & \textbf{P} & \textbf{R}\\
\hline
\bert & \asd{71.03}{1.12} & \asd{65.24}{1.12} & \asd{77.95}{1.56} & \asd{62.94}{1.39} & \asd{57.89}{1.29} & \asd{69.00}{2.44}\\
\distilbert & \asd{69.53}{0.81} & \asd{67.45}{2.42} & \asd{71.86}{1.30} & \asd{60.38}{1.30} & \asd{58.58}{2.65} & \asd{62.39}{0.90}\\
\spanbert & \asd{60.75}{0.07} & \asd{58.66}{1.26} & \asd{63.04}{1.31} & \asd{50.06}{0.68} & \asd{47.18}{0.82} & \asd{53.33}{0.49}\\
\roberta & \asd{61.56}{3.87} & \asd{59.17}{4.12} & \asd{64.18}{3.72} & \asd{50.23}{6.63} & \asd{48.30}{6.64} & \asd{52.35}{6.67}\\
\electra & \asd{71.36}{1.49} & \asd{68.22}{2.20} & \asd{74.85}{1.38} & \asd{62.64}{1.51} & \asd{59.89}{2.22} & \asd{65.69}{1.10}\\
\xlnet & \asd{72.90}{1.28} & \asd{69.26}{1.50} & \asd{76.97}{1.46} & \asd{64.35}{1.90} & \asd{61.14}{2.08} & \asd{67.94}{1.90}\\
\hline
\bertweet & \asd{73.04}{1.06} & \asd{70.68}{1.29} & \asd{75.59}{1.65} & \asd{63.99}{1.09} & \asd{62.44}{0.63} & \asd{65.67}{2.11}\\
\biobert & \asd{68.53}{0.91} & \asd{65.44}{0.68} & \asd{71.96}{2.00} & \asd{60.83}{1.41} & \asd{58.08}{0.57} & \asd{63.88}{2.52}\\
\bioclinicalbert & \asd{67.16}{1.21} & \asd{61.84}{1.12} & \asd{73.50}{1.71} & \asd{57.54}{1.45} & \asd{52.98}{1.27} & \asd{62.97}{1.92}\\
\scibert & \asd{66.09}{2.07} & \asd{61.91}{1.89} & \asd{70.94}{3.03} & \asd{57.75}{2.64} & \asd{53.59}{2.95} & \asd{62.64}{2.37}\\
\pubmedbert & \asd{64.21}{4.46} & \asd{62.86}{4.58} & \asd{65.75}{5.39} & \asd{52.86}{7.46} & \asd{52.67}{8.09} & \asd{53.09}{6.89}\\
\xyendr & \asd{72.95}{1.47} & \asd{69.45}{1.34} & \asd{76.87}{2.37} & \asd{65.24}{2.34} & \asd{62.09}{1.66} & \asd{68.77}{3.38}\\
\bioroberta & \asd{62.20}{2.10} & \asd{57.06}{3.04} & \asd{68.44}{1.87} & \asd{51.91}{2.92} & \asd{47.70}{3.34} & \asd{57.03}{3.04}\\
\bioelectra & \asd{68.97}{1.49} & \asd{67.15}{2.03} & \asd{70.94}{1.58} & \asd{60.56}{1.96} & \asd{58.91}{2.53} & \asd{62.34}{1.77}\\
\hline
\end{tabular}
}
\caption{Metrics of the \autoencoder models with LSTM module on SMM4H.}
\label{tab:bert_lstm_smm4h20}
\end{table*}

\begin{table*}[!hbtp]
\centering
\resizebox{\linewidth}{!}{%

\begin{tabular}{r ccc ccc}
\hline
& \multicolumn{3}{c}{\textbf{Relaxed}} & \multicolumn{3}{c}{\textbf{Strict}}\\
& \textbf{F1} & \textbf{P} & \textbf{R} & \textbf{F1} & \textbf{P} & \textbf{R}\\
\hline
\bert & \asd{78.76}{0.35} & \asd{76.92}{0.73} & \asd{80.71}{0.93} & \asd{66.67}{0.40} & \asd{65.11}{0.58} & \asd{68.32}{0.91}\\
\distilbert & \asd{76.30}{0.34} & \asd{73.29}{0.25} & \asd{79.57}{0.53} & \asd{62.90}{0.71} & \asd{60.31}{0.74} & \asd{65.73}{0.81}\\
\spanbert & \asd{79.58}{0.20} & \asd{79.62}{0.82} & \asd{79.54}{0.44} & \asd{68.12}{0.30} & \asd{68.14}{0.71} & \asd{68.10}{0.58}\\
\roberta & \asd{78.31}{0.32} & \asd{77.08}{0.59} & \asd{79.58}{0.60} & \asd{65.83}{0.45} & \asd{64.79}{0.50} & \asd{66.90}{0.71}\\
\electra & \asd{79.30}{0.27} & \asd{77.45}{0.80} & \asd{81.25}{0.47} & \asd{67.34}{0.70} & \asd{65.87}{0.58} & \asd{68.89}{1.27}\\
\xlnet & \asd{79.35}{0.64} & \asd{77.63}{1.04} & \asd{81.15}{0.30} & \asd{67.48}{0.74} & \asd{66.05}{1.07} & \asd{68.97}{0.49}\\
\gpttwo & \asd{27.55}{5.93} & \asd{28.69}{5.24} & \asd{26.8}{7.03} & \asd{12.98}{3.91} & \asd{13.47}{3.53} & \asd{12.67}{4.38}\\
\tfive & \asd{62.77}{0.62} & \asd{53.15}{0.82} & \asd{76.65}{0.59} & \asd{52.98}{0.88} & \asd{44.86}{0.92} & \asd{64.7}{0.98}\\
\pegasus & \asd{61.31}{0.65} & \asd{51.49}{1.14} & \asd{75.78}{0.72} & \asd{50.51}{0.76} & \asd{42.42}{0.97} & \asd{62.43}{1.11}\\
\bart & \asd{57.98}{0.64} & \asd{47.69}{0.85} & \asd{73.95}{0.56} & \asd{47.40}{0.78} & \asd{38.99}{0.84} & \asd{60.45}{0.87}\\
\hline
\bertweet & \asd{78.28}{0.47} & \asd{75.93}{0.36} & \asd{80.78}{0.78} & \asd{65.51}{1.01} & \asd{63.43}{0.92} & \asd{67.73}{1.12}\\
\biobert & \asd{78.32}{0.43} & \asd{77.90}{0.84} & \asd{78.75}{0.39} & \asd{65.97}{0.75} & \asd{65.62}{1.03} & \asd{66.33}{0.64}\\
\bioclinicalbert & \asd{78.09}{0.28} & \asd{77.07}{0.91} & \asd{79.17}{1.14} & \asd{66.23}{0.58} & \asd{65.09}{0.98} & \asd{67.44}{0.95}\\
\scibert & \asd{79.22}{0.42} & \asd{77.77}{0.69} & \asd{80.73}{0.37} & \asd{67.63}{0.63} & \asd{66.44}{0.78} & \asd{68.88}{0.60}\\
\pubmedbert & \asd{79.18}{0.55} & \asd{77.67}{0.84} & \asd{80.76}{0.41} & \asd{67.16}{0.80} & \asd{65.85}{1.07} & \asd{68.51}{0.63}\\
\xyendr & \asd{80.57}{0.45} & \asd{79.08}{0.94} & \asd{82.12}{0.40} & \asd{69.12}{0.66} & \asd{67.62}{1.20} & \asd{70.69}{0.28}\\
\bioroberta & \asd{77.77}{0.23} & \asd{76.16}{0.81} & \asd{79.48}{1.23} & \asd{65.53}{0.37} & \asd{63.83}{0.63} & \asd{67.34}{0.88}\\
\bioelectra & \asd{78.25}{0.53} & \asd{76.92}{0.97} & \asd{79.66}{0.96} & \asd{66.20}{0.82} & \asd{65.01}{1.06} & \asd{67.45}{1.09}\\
\scifive & \asd{62.80}{0.19} & \asd{53.09}{0.25} & \asd{76.84}{0.36} & \asd{52.74}{0.53} & \asd{44.59}{0.46} & \asd{64.53}{0.72}\\
\hline
\end{tabular}
}
\caption{Metrics of all the base models on CADEC.}
\label{tab:bert_wrapper_cadec}
\end{table*}

\begin{table*}[!hbtp]
\centering
\resizebox{\linewidth}{!}{%
\begin{tabular}{r ccc ccc}
\hline
& \multicolumn{3}{c}{\textbf{Relaxed}} & \multicolumn{3}{c}{\textbf{Strict}}\\
& \textbf{F1} & \textbf{P} & \textbf{R} & \textbf{F1} & \textbf{P} & \textbf{R}\\
\hline
\bert & \asd{78.89}{0.55} & \asd{76.87}{0.54} & \asd{81.01}{0.88} & \asd{66.86}{0.86} & \asd{65.12}{0.69} & \asd{68.70}{1.18}\\
\distilbert & \asd{77.92}{0.25} & \asd{75.97}{0.28} & \asd{79.97}{0.68} & \asd{65.40}{0.53} & \asd{63.75}{0.29} & \asd{67.15}{0.90}\\
\spanbert & \asd{79.36}{0.20} & \asd{78.89}{0.43} & \asd{79.83}{0.30} & \asd{67.72}{0.40} & \asd{67.15}{0.66} & \asd{68.30}{0.25}\\
\roberta & \asd{78.50}{0.52} & \asd{77.63}{0.94} & \asd{79.39}{0.44} & \asd{66.08}{1.03} & \asd{65.58}{1.11} & \asd{66.60}{1.26}\\
\electra & \asd{79.67}{0.62} & \asd{77.15}{0.85} & \asd{82.36}{0.66} & \asd{67.82}{0.75} & \asd{65.48}{1.03} & \asd{70.35}{0.73}\\
\xlnet & \asd{79.44}{0.27} & \asd{77.37}{0.32} & \asd{81.63}{0.73} & \asd{67.98}{0.33} & \asd{66.07}{0.43} & \asd{70.02}{0.65}\\
\hline
\bertweet & \asd{79.75}{0.25} & \asd{78.88}{0.50} & \asd{80.63}{0.13} & \asd{68.45}{0.37} & \asd{67.61}{0.72} & \asd{69.31}{0.25}\\
\biobert & \asd{77.55}{0.27} & \asd{76.32}{0.43} & \asd{78.83}{0.35} & \asd{64.54}{0.53} & \asd{63.63}{0.64} & \asd{65.49}{0.53}\\
\bioclinicalbert & \asd{78.49}{0.24} & \asd{76.44}{0.92} & \asd{80.67}{0.54} & \asd{66.78}{0.34} & \asd{65.07}{0.89} & \asd{68.60}{0.32}\\
\scibert & \asd{78.32}{0.42} & \asd{76.67}{0.91} & \asd{80.04}{0.25} & \asd{65.99}{0.74} & \asd{64.61}{1.12} & \asd{67.44}{0.42}\\
\pubmedbert & \asd{78.98}{0.36} & \asd{77.71}{0.52} & \asd{80.28}{0.48} & \asd{67.29}{0.26} & \asd{66.14}{0.91} & \asd{68.49}{0.48}\\
\xyendr & \asd{79.64}{0.64} & \asd{77.98}{1.20} & \asd{81.40}{0.32} & \asd{67.82}{1.21} & \asd{66.40}{1.64} & \asd{69.31}{0.82}\\
\bioroberta & \asd{77.45}{0.33} & \asd{75.53}{0.32} & \asd{79.47}{0.74} & \asd{64.53}{0.75} & \asd{62.71}{0.37} & \asd{66.46}{1.19}\\
\bioelectra & \asd{79.22}{0.35} & \asd{77.33}{0.35} & \asd{81.21}{0.82} & \asd{67.82}{0.65} & \asd{66.10}{0.51} & \asd{69.64}{1.01}\\
\hline

\end{tabular}
}
\caption{Metrics of the \autoencoder models with CRF module on CADEC.}
\label{tab:bert_crf_cadec}
\end{table*}

\begin{table*}[!hbtp]
\centering
\resizebox{\linewidth}{!}{%
\begin{tabular}{r ccc ccc}
\hline
& \multicolumn{3}{c}{\textbf{Relaxed}} & \multicolumn{3}{c}{\textbf{Strict}}\\
& \textbf{F1} & \textbf{P} & \textbf{R} & \textbf{F1} & \textbf{P} & \textbf{R}\\
\hline
\bert & \asd{78.89}{0.30} & \asd{77.86}{0.74} & \asd{79.95}{0.66} & \asd{66.95}{0.48} & \asd{65.78}{0.51} & \asd{68.16}{0.75}\\
\distilbert & \asd{77.84}{0.40} & \asd{76.75}{1.03} & \asd{78.99}{1.22} & \asd{65.73}{0.42} & \asd{64.84}{0.71} & \asd{66.67}{1.19}\\
\spanbert & \asd{78.48}{0.53} & \asd{78.22}{0.66} & \asd{78.74}{1.06} & \asd{66.25}{1.09} & \asd{66.04}{1.20} & \asd{66.48}{1.28}\\
\roberta & \asd{78.01}{0.46} & \asd{77.06}{0.93} & \asd{79.00}{0.65} & \asd{65.56}{0.50} & \asd{64.61}{0.59} & \asd{66.54}{0.75}\\
\electra & \asd{79.30}{0.47} & \asd{77.40}{1.22} & \asd{81.32}{0.41} & \asd{66.96}{0.65} & \asd{65.35}{1.26} & \asd{68.66}{0.30}\\
\xlnet & \asd{79.00}{0.35} & \asd{77.95}{0.78} & \asd{80.08}{0.47} & \asd{66.95}{0.53} & \asd{66.17}{0.83} & \asd{67.75}{0.27}\\
\hline
\bertweet & \asd{78.66}{1.85} & \asd{78.02}{2.67} & \asd{79.32}{1.23} & \asd{66.72}{2.87} & \asd{66.19}{3.49} & \asd{67.27}{2.31}\\
\biobert & \asd{78.63}{0.27} & \asd{78.34}{0.51} & \asd{78.92}{0.41} & \asd{66.62}{0.34} & \asd{66.40}{0.34} & \asd{66.85}{0.59}\\
\bioclinicalbert & \asd{78.79}{0.37} & \asd{77.02}{0.65} & \asd{80.65}{0.37} & \asd{67.49}{0.58} & \asd{65.97}{0.81} & \asd{69.08}{0.46}\\
\scibert & \asd{79.31}{0.32} & \asd{77.32}{0.23} & \asd{81.41}{0.51} & \asd{67.80}{0.69} & \asd{66.00}{0.94} & \asd{69.72}{0.88}\\
\pubmedbert & \asd{78.50}{0.71} & \asd{78.30}{1.66} & \asd{78.73}{0.87} & \asd{66.73}{1.16} & \asd{66.72}{1.98} & \asd{66.76}{0.86}\\
\xyendr & \asd{75.27}{4.90} & \asd{73.12}{3.82} & \asd{77.62}{6.27} & \asd{62.21}{6.61} & \asd{60.39}{5.64} & \asd{64.19}{7.75}\\
\bioroberta & \asd{78.09}{0.30} & \asd{77.55}{0.86} & \asd{78.65}{0.93} & \asd{65.54}{0.63} & \asd{64.83}{0.38} & \asd{66.27}{1.20}\\
\bioelectra & \asd{79.33}{0.31} & \asd{78.41}{0.45} & \asd{80.28}{0.58} & \asd{67.89}{0.65} & \asd{67.10}{0.60} & \asd{68.70}{0.86}\\
\hline
\end{tabular}
}
\caption{Metrics of the \autoencoder models with LSTM module on CADEC.}
\label{tab:bert_lstm_cadec}
\end{table*}

\clearpage
\newpage

 \bibliographystyle{elsarticle-num} 
 \bibliography{cas-refs}





\end{document}